\documentclass[acmsmall]{acmart}
\usepackage{amsmath,amsfonts}
\usepackage{mathrsfs}
\usepackage{graphicx}
\usepackage{textcomp}
\usepackage{xcolor}
\usepackage{color}
\usepackage{subfigure}
\usepackage{booktabs}
\usepackage{multirow}
\usepackage{makecell}
\usepackage{verbatim}
\usepackage{enumitem}
\usepackage{flushend}
\usepackage{bm}
\usepackage{bbm}
\usepackage{hyperref}
\usepackage{algorithmic}
\usepackage{array}
\usepackage{url}
\usepackage{todonotes}
\usepackage{algorithm}
\usepackage{algorithmic}

\usepackage{pifont}

\newcommand{\xmark}{\ding{55}}

\AtBeginDocument{%
  \providecommand\BibTeX{{%
    \normalfont B\kern-0.5em{\scshape i\kern-0.25em b}\kern-0.8em\TeX}}}

\setcopyright{acmcopyright}
\copyrightyear{2022}
\acmYear{2022}
\acmDOI{XXXXXXX.XXXXXXX}

\begin{document}

\title{Geometry Contrastive Learning on Heterogeneous Graphs}

\author{Shichao Zhu}
\email{zhushichao@iie.ac.cn}
\affiliation{%
  \institution{Institute of Information Engineering, Chinese Academy of Sciences. School of Cyber Security, University of Chinese Academy of Sciences}
  \city{Beijing}
  \country{China}
}

\author{Chuan Zhou}
\authornote{Corresponding author.}
\email{zhouchuan@amss.ac.cn}
\affiliation{%
  \institution{Academy of Mathematics and Systems Science, Chinese Academy of Sciences. School of Cyber Security, University of Chinese Academy of Sciences}
  \city{Beijing}
  \country{China}
 }

\author{Anfeng Cheng}
\email{chenganfeng01@baidu.com}
\affiliation{%
  \institution{Baidu Inc.}
  \city{Beijing}
  \country{China}
}

\author{Shirui Pan}
\email{shiruipan@ieee.org}
\affiliation{
  \institution{Griffith University}
  \city{Queensland}
  \country{Australia}
 }
 
\author{Shuaiqiang Wang, Dawei Yin}
\email{wangshuaiqiang@baidu.com, yindawei@acm.org}
\affiliation{%
  \institution{Baidu Inc.}
  \city{Beijing}
  \country{China}
}

\author{Bin Wang}
\email{wangbin11@xiaomi.com}
\affiliation{%
  \institution{Xiaomi AI Lab}
  \city{Beijing}
  \country{China}
}

\renewcommand{\shortauthors}{Shichao, et al.}

\begin{abstract}
Self-supervised learning (especially contrastive learning) methods on heterogeneous graphs can effectively get rid of the dependence on supervisory data. Meanwhile, most existing representation learning methods embed the heterogeneous graphs into a single geometric space, either Euclidean or hyperbolic. This kind of single geometric view is usually not enough to observe the complete picture of heterogeneous graphs due to their rich semantics and complex structures. Under these observations, this paper proposes a novel self-supervised learning method, termed as Geometry Contrastive Learning (GCL), to better represent the heterogeneous graphs when supervisory data is unavailable. GCL views a heterogeneous graph from Euclidean and hyperbolic perspective simultaneously, aiming to make a strong merger of the ability of modeling rich semantics and complex structures, which is expected to bring in more benefits for downstream tasks. GCL maximizes the mutual information between two geometric views by contrasting representations at both local-local and local-global semantic levels. Extensive experiments on four benchmarks data sets show that the proposed approach outperforms the strong baselines, including both unsupervised methods and supervised methods, on three tasks—node classification, node clustering and similarity search.

\end{abstract}



\maketitle
\section{Introduction}
In reality, with multiple types of nodes and edges, heterogeneous graphs \cite{sun2013mining,shi2016survey,ji2021survey} are ubiquitous in characterizing semantic relations between objects, ranging from biological networks \cite{gilmer2017neural} to social networks \cite{shi2018heterogeneous,fan2019metapath}. 
It is becoming more and more important to effectively mine and learn from heterogeneous graphs.
Graph Neural Networks (GNNs) \cite{kipf2016semi} in particular are effective techniques to learn graphs directly from the structural level for analyzing the underlying symbolic nature of graphs \cite{xu2018powerful,jiao2021temporal}.
Modeling heterogeneous graphs with GNNs has achieved state-of-the-art results, such as R-GCN \cite{schlichtkrull2018modeling} and HAN \cite{wang2019han}. 
GNNs typically require task-dependent labels to capture the inherent information of graph to learn representations. However, labeling graph is costly due to the requirements of prior multi-domain knowledge, which leads to the label scarcity issue especially on heterogeneous graphs. 
Some self-supervised methods coupled with heterogeneous GNNs are therefore introduced to get rid of the dependence of supervisory data, in which contrastive learning has raised a recent surge of interest and achieved state-of-the-art results. For example, DMGI \cite{park2020unsupervised} and HDGI \cite{ren2019heterogeneous} maximize the mutual information (MI) between local representation and global representation by contrastive learning to learn the heterogeneous graph encoder.

Typically, there are two prominent characteristics on heterogeneous graphs that make the representation learning challenging. 1) \emph{Rich semantics}. There exist multiple types of nodes and edges intersecting between each other, leading to a large number of semantic correlations, which are yet to be fully excavated \cite{park2020unsupervised,ren2019heterogeneous}. 2) \emph{Complex structures}. Many real-word heterogeneous graphs exhibit a non-Euclidean geometry such as scale-free or hierarchical structure \cite{clauset2008hierarchical,bronstein2017geometric}. 
These characteristics lead to an interesting and fundamental question that, \emph{which kind of geometric space, as the latent feature space, should be employed to better express the rich semantic and complex structure of heterogeneous graphs?} Most current GNNs \cite{wang2019han,ren2019heterogeneous,hu2020heterogeneous} select to embed heterogeneous graphs into the Euclidean space, due to its intuition-friendly generalization with powerful simplicity and efficiency. Some other works \cite{wang2019hyperbolic,ganea2018hyperbolic,chami2019hgcn} believe that hyperbolic space has more advantages in expressing heterogeneous graphs, especially in describing the hierarchical structure. 
No matter which geometric space is taken, these works assume that all the nodes in a heterogeneous graph share the same spatial curvature and embed graph into a single geometric space, either Euclidean or hyperbolic. This kind of single geometric view is usually not enough to effectively observe the complete picture of complex heterogeneous graphs. As described in work \cite{zhu2020gil}, Euclidean space is suitable for representing the nodes with low-dimensional regular structure, while nodes with hierarchical structure are more suitable for hyperbolic space. Due to the rich semantic and complex structural information in heterogeneous graphs, learning embedding in a single space cannot usually guarantee the optimality of all nodes' representation, which will further inevitably affect the effect of self-supervised learning on heterogeneous graphs.


\par
To address above issues and facilitate a more comprehensive representation for heterogeneous graphs, we introduce a novel geometry contrastive learning (GCL) for heterogeneous graphs, which exploits hyperbolic and Euclidean geometries jointly for heterogeneous graph embedding under the framework of contrastive learning to capture rich semantic and complex structural information in heterogeneous graphs.
GCL not only considers how to model different augmented graphs in Euclidean and hyperbolic geometric spaces, but also promotes the maximization of mutual information (MI) between the latent embeddings at local-local and local-global semantic levels from two different geometric views, 
which provides more expressive power for inherent complex structure with robustness to the geometries. 
Specifically, two dedicated geometric encoders are introduced for each augmented graph, to learn different geometric representations. 
The whole model is trained end-to-end, using a composition of contrastive losses to maximize the MI between latent embeddings from two graph geometric views at local-local and local-global semantic levels. 
The heterogeneity of the graph is encoded in both graph encoders and MI measurements in the framework of contrastive learning. 
In particular, the relation-aware one-hop and multi-hop topologies are introduced as the local and global semantic levels, respectively, in order to enhance the capability of capturing the rich semantic correlations on heterogeneous graphs. 

\par
Our extensive evaluations demonstrate that our method outperforms SOTA methods in node classification, node clustering and similarity search tasks on four benchmark data sets. Furthermore, we systematically study the major components of our framework, including graph geometric views and contrastive modes. 
The major contributions of this paper are summarized as follows. 
\begin{itemize}
\item To our best knowledge, we are the first to study the problem of geometry contrastive learning on heterogeneous graphs, which unifies the Euclidean and hyperbolic geometric views to better express the heterogeneity.
\item We propose a novel geometry contrastive learning (GCL) framework to deal with rich semantics and complex structure simultaneously, to maximize the agreement of both local-local and local-global representations in different geometric spaces.
\item Extensive experimental results have verified the effectiveness of our method in terms of node classification, node clustering and similarity search on benchmark data sets. The analysis further validates the rationality of GCL in terms of graph geometric views and contrastive modes.
\end{itemize}
\par
The remainder of this paper is organized as follows. In Section \ref{sec:related}, we first review the related works. Then, the preliminary notations and definitions are introduced in Section \ref{sec:pre}. In Section \ref{sec:model}, we introduce the overall pipeline and the components of our framework in detail. After that, Section \ref{sec:exp} illustrates the experimental results and analysis. Finally, Section \ref{sec:conclusion} concludes the paper and forecasts the future research directions.

\section{Related Works}\label{sec:related}
\subsection{Heterogeneous Graph Embedding}
Heterogeneous graph embedding is proposed to embed heterogeneous graphs into a low dimensional space while preserving the structure and property. And then the learned embedding can be applied to downstream tasks. The existing methods can be roughly divided into shallow models \cite{jiang2017semi,dong2017metapath2vec,fu2017hin2vec} and deep models \cite{chang2015heterogeneous,wang2019han,schlichtkrull2018modeling}. The shallow models include random walk based network embedding \cite{grover2016node2vec,jiang2017semi}, meta-path based context embedding \cite{dong2017metapath2vec,fu2017hin2vec}, hyperbolic space based embedding \cite{wang2019hyperbolic}, etc. The deep models mainly focus on GNN-based manner \cite{chang2015heterogeneous} such as HAN \cite{wang2019han}, R-GCN \cite{schlichtkrull2018modeling} and mGCN \cite{ma2019multi}. HAN considers the attention mechanism based on meta-paths for heterogeneous graph learning. R-GCN splits heterogeneous graph to multiple homogeneous subgraphs by building an independent adjacency matrix for each type of edge, and then applies GCN \cite{kipf2016semi} on each subgraph. 
Besides, HetGNN \cite{zhang2019hetgnn} uses bi-LSTM to aggregate
the embedding of those sampled neighboring nodes so as to learn the deep interactions among heterogeneous nodes. HGT \cite{hu2019hgt} presents the heterogeneous graph transformer architecture for modeling Web-scale heterogeneous graphs with corresponding mini-batch sampling algorithm.
GTN \cite{yun2019gtn} designs an aggregation function, which can find the suitable metapaths automatically during the process of message passing.
The above heterogeneous GNNs are mostly semi-supervised or supervised methods, which relies on task-dependent labels. Recent works \cite{hwang2020self} show that the representation can be further improved by self-supervised signals by predicting meta-paths as auxiliary tasks for heterogeneous graph.

\subsection{Graph Geometry Learning}
Non-Euclidean Riemannian spaces have recently gained extensive attention in learning representation for non-euclidean data. \cite{nickel2017poincare} was the first work to learn hierarchical embeddings in hyperbolic space for link prediction. Following this work, \cite{balazevic2019multi} proposes a translation model in hyperbolic space for multi-relational graph embedding. HHNE \citep{wang2019hyperbolic} conducts the meta-path guided random walk in hyperbolic spaces, where the similarity between nodes can be measured as the hyperbolic distance. In the hyperbolic space, some properties of heterogeneous graphs, e.g., hierarchical and power-law structures, can be naturally reflected in the learned node embeddings.
Going beyond GCNs in Euclidean space \citep{bruna2013spectral, kipf2016semi}, HGNN \citep{liu2019hgnn} and HGCN \citep{chami2019hgcn} first extend GCNs into hyperbolic space, achieving state-of-the-art performance on learning graph embedding for scale-free networks. 
To model graphs of mixed topologies, $\kappa$-GCN \citep{bachmann2020constant} unifies curvatures in a $\kappa$-stereographic and then extends GCNs into the products of Riemannian projection manifolds. And GIL \citep{zhu2020gil} proposes graph geometric interaction learning that models interaction between hyperbolic and Euclidean spaces. 

\subsection{Graph Contrastive Learning}
The contrastive learning on graphs has received a lot of attention recently, and has achieved state-of-the-art performance in self-supervised graph representation learning, such as node classification and graph classification tasks \cite{qiu2020gcc,sun2021mocl,liu2021anomaly}.
The main idea of contrastive learning is to measure the loss in latent space by contrasting samples from a distribution that contains correlations of interest and the distribution that does not. 
Typically, deep graph infomax (DGI) \cite{velickovic2019deep} extends the deep InfoMax \cite{hjelm2018learning} to graphs, which encourages an encoder to learn representations through maximizing the MI between local representation and global representation.
Besides, a rising sub-domain that extends contrastive learning into multiple views and has achieved impressive performance on downstream tasks.
Specifically, contrastive multi-view learning has recently applied in visual representation learning \cite{chen2020simple,tian2019contrastive} and homogeneous graph representation learning \cite{hassani2020contrastive,zhu2020deep}.
The two representative works MVGRL \cite{hassani2020contrastive} and GRACE \cite{zhu2020deep} extend the InfoMax principle \cite{linsker1988self} to multiple views and maximize the MI across views generated by composition of data augmentations on homogeneous graphs. They provide multi-view for graph learning via introducing an additional diffusion matrix and data augmentations, respectively. 

Following the same spirit that maximizes MI between local and global representations, HDGI \cite{ren2019heterogeneous} and DMGI \cite{park2020unsupervised} extend contrastive learning to heterogeneous graphs. HDGI \cite{ren2019heterogeneous} exploits HAN\cite{wang2019han} as graph encoder to obtain node embeddings and then introduces MI maximization as the training objective. 
To minimize the disagreements among the relation-type specific node embeddings, DMGI \cite{park2020unsupervised} introduces an extra consensus embedding to integrate node embeddings from multiple types as a regularization module to serve the major MI objective. 
PT-HGNN \cite{jiang2021pthgnn} is a pre-training framework via contrastive learning for large-scale heterogeneous graphs, which proposes both node and schema-level pre-training tasks to utilize node relations and the network schema, respectively.
The above contrastive learning on heterogeneous graphs contrast embeddings from one view learned by the shared encoder. 
A recent work HeCo \cite{wang2021cocontrastive} utilizes both network schema and meta-path as two structural views for co-contrastive learning, to capture both local structure and high-order structure simultaneously.

\par
Different from the above methods, our method first introduces the graph geometry learning into the framework of contrastive learning to better capture the heterogeneity existing in graphs. Furthermore, we systematically study the effectiveness of the contrastive modes and graph geometric views. 


\section{Notations and Preliminaries}\label{sec:pre}
We here introduce some basic concepts and formalize the problem of graph embedding. For easy retrieval, we summarize the commonly used notations in Table \ref{tab:notations}.

\begin{table}[t!]
\caption{Commonly used notations. The three blocks of the table show the notation of variables about heterogeneous graphs, multi-view contrastive learning and graph encoders.
}
\label{tab:notations}
\centering
\begin{tabular}{cl}
\toprule
 Notations & Descriptions \\
\midrule
  $G=(V,E,T,R)$ & A heterogeneous graph \\
  $V$ & The node set of graph $G$ \\
  $E$ & The edge set of graph $G$ \\
  $T$ & The node types set of graph $G$ \\
  $R$ & The relation set of graph $G$ \\
  $N$ & The number of nodes in graph $G$ \\
  $A\in\mathbb{R}^{N\times N}$ & The adjacency matrix of graph $G$ \\
  $X\in\mathbb{R}^{N\times F}$ & The node attributes of graph $G$\\
  $G^{m}$ & The meta-path $m$ specific subgraph in $G$ \\
  $\phi_n(\cdot)$ & The node type mapping function $\phi_n:V\to T$ \\
  $\phi_e(\cdot)$ & The edge type mapping function $\phi_e:E\to R$ \\
  \midrule
  \multirow{2}{*}{$\mathcal{G}_1=(X_1, A_1)$} & A graph view generated by data augmentation \\
  ~ & as input of Euclidean graph encoder \\
  \multirow{2}{*}{$\mathcal{G}_2=(X_2, A_2)$} & A graph view generated by data augmentation \\
  ~ & as input of hyperbolic graph encoder \\
  $f_{\theta}(\cdot)$ & Euclidean graph encoder $f_{\theta}: V\to\mathbb{R}^d$ \\
  $f_{\omega}(\cdot)$ & hyperbolic graph encoder $f_{\omega}: V\to\mathbb{D}^d$ \\
  $f_{\phi}(\cdot)$ & The shared projection head $f_{\phi}: \mathbb{R}^d\to\mathbb{R}^d$ \\
  \multirow{2}{*}{$\bm{h}_i^{m}$} & The meta-path $m$ specific hidden representation  \\
  ~ & vector of node $i$ \\
  \multirow{2}{*}{$\bm{h}_g^{m}$} & The meta-path $m$ specific hidden representation \\ 
  ~ & vector of subgraph $G^{m}$ \\\
  \multirow{2}{*}{$\bm{h}_i$, $\widetilde{\bm{h}}_i$} & The integrated hidden representation vector of\\
  ~ & node $i$ in Euclidean and hyperbolic spaces\\
  \multirow{2}{*}{$\bm{h}_g$, $\widetilde{\bm{h}}_g$} & The integrated hidden representation vector of  \\
  ~ & graph $G$ in Euclidean and hyperbolic spaces \\
  $\bm{H}^{m}\in\mathbb{R}^{N\times d}$ & The learned node embedding matrix of subgraph $G^m$\\
  $\bm{H}\in\mathbb{R}^{N\times d}$ & The learned node embedding matrix of graph $G$\\
  \midrule
  \multirow{2}{*}{$\langle u,r,v \rangle$} & A positive triple in graph $G$ that satisfies  \\ 
  ~ & $(u,v)\in E, \phi_e(u,v)=r, r\in R$\\
  \multirow{2}{*}{$\mathcal{N}^{node}_{\langle u,v \rangle}$} & The set of negative samples with unrelated   \\ 
  ~ & nodes $(u,v^{-})\notin E$ \\
  \multirow{2}{*}{$\mathcal{N}^{rel}_{\langle u,r,v \rangle}$} & The set of negative samples with inconsistent  \\ 
  ~ & relations $\langle u,r^{-},v\rangle, r^{-}\in R\setminus r$\\
\bottomrule
\end{tabular}
\end{table}
\par
\vspace{2mm}
\noindent\textbf{\textit{Definition 1. Heterogeneous Graph.}}\quad A heterogeneous graph \cite{shi2016survey,wang2020survey} $G=(V,E,T,R)$ is a graph with a set of nodes $V$, edges $E$, node types $T$ and edge types (or relation types) $R$, where the number of node types $|T|$ and edge types $|R|$ satisfy $|T|+|R|>2$. 
It is also associated with a node type mapping function $\phi_n:V\rightarrow T$ and an edge type mapping function $\phi_e:E\rightarrow R$.
When $|T|=1$ and $|R|=1$, it becomes a homogeneous graph.
The node attributes can be encoded as initial feature matrix $X\in\mathbb{R}^{N\times F}$, where $N$ and $F$ denote the number of nodes and dimension of initial node features.
\vspace{2mm}

\noindent\textbf{\textit{Definition 2. Meta-Path based Subgraph.}}\quad A meta-path \cite{sun2011pathsim} $m$ can be viewed as a composite relation, consisting of a sequence of relations. 
The nodes can be connected by a meta-path, which is denoted as $m=v_1\overset{r_1}{\rightarrow}v_2\overset{r_2}{\rightarrow}...\overset{r_l}{\rightarrow}v_{l+1}$, where $r_i\in R$ represents the $i$-th relation type of meta-path $m$. 
Based on each meta-path $m$, we can extract a subgraph $G^{m}$, in which nodes are connected by the meta-path $m$. We call it as a meta-path based subgraph. The corresponding adjacency matrix for $G^{m}$ is defined as $A^{m}\in\mathbb{R}^{N\times N}$, where $A_{ij}^m=1$ if node $v_i$ and $v_j$ are connected by meta-path $m$.
\par
The meta-path based subgraph is defined to extract rich semantics including multi-hop connections, and has been shown to be useful to analyze heterogeneous graphs with few edge types \cite{shi2018heterogeneous,ren2019heterogeneous}. For instance, Paper-Author-Paper (PAP) and Paper-Subject-Paper (PSP) are two typical meta-paths, which contain the semantic "papers written by the same author" and "papers belonging to the same subject", respectively.
\par
\vspace{2mm}
\noindent\textbf{\textit{Definition 3. Heterogeneous Graph Embedding.}}\quad Given a heterogeneous graph $G=(V,E,T,R)$, the goal of heterogeneous graph embedding is to learn a node mapping function that projects each node into a low dimensional space, while preserving the graph structure and semantic correlations. The node representation can be embedded into Euclidean space with $f_{\theta}: V\to \mathbb{R}^d$, where $d\ll \left|V\right|$. 
The node representation can also be embedded into hyperbolic space with low dimensions, where the mapping function can be defined as $f_{\omega}: V\to \mathbb{D}^d$. 
The final learned node representation $\bm{H}\in\mathbb{R}^{N\times d}$ is applied to downstream tasks such as node classification and node clustering. In addition, the node representation in a meta-path based subgraph $G^{m}$ can be defined as $\bm{H}^{m}$.
\par
\vspace{2mm}
\noindent\textbf{\textit{Definition 4. Contrastive Learning.}}\quad The goal of contrastive learning is to learn an embedding by separating samples from different distributions via a contrastive loss $\mathcal{L}$. The objective $\mathcal{L}$ is a function that aims to reflect incompatibility of each $(\bm{q},\bm{k})$ pair, where $\bm{q},\bm{k}\in\mathbb{R}^d$ denote the \textit{query} and \textit{key} vectors with $d$ dimensional embeddings. The key vector set consists of positive and negative samples from different distributions. An ideal objective $\mathcal{L}$ would bring the positive sample pairs closer and the negative sample pairs farther apart.

\section{Methodology}\label{sec:model}
\begin{figure*}[t]
\centering
\includegraphics[scale=0.30]{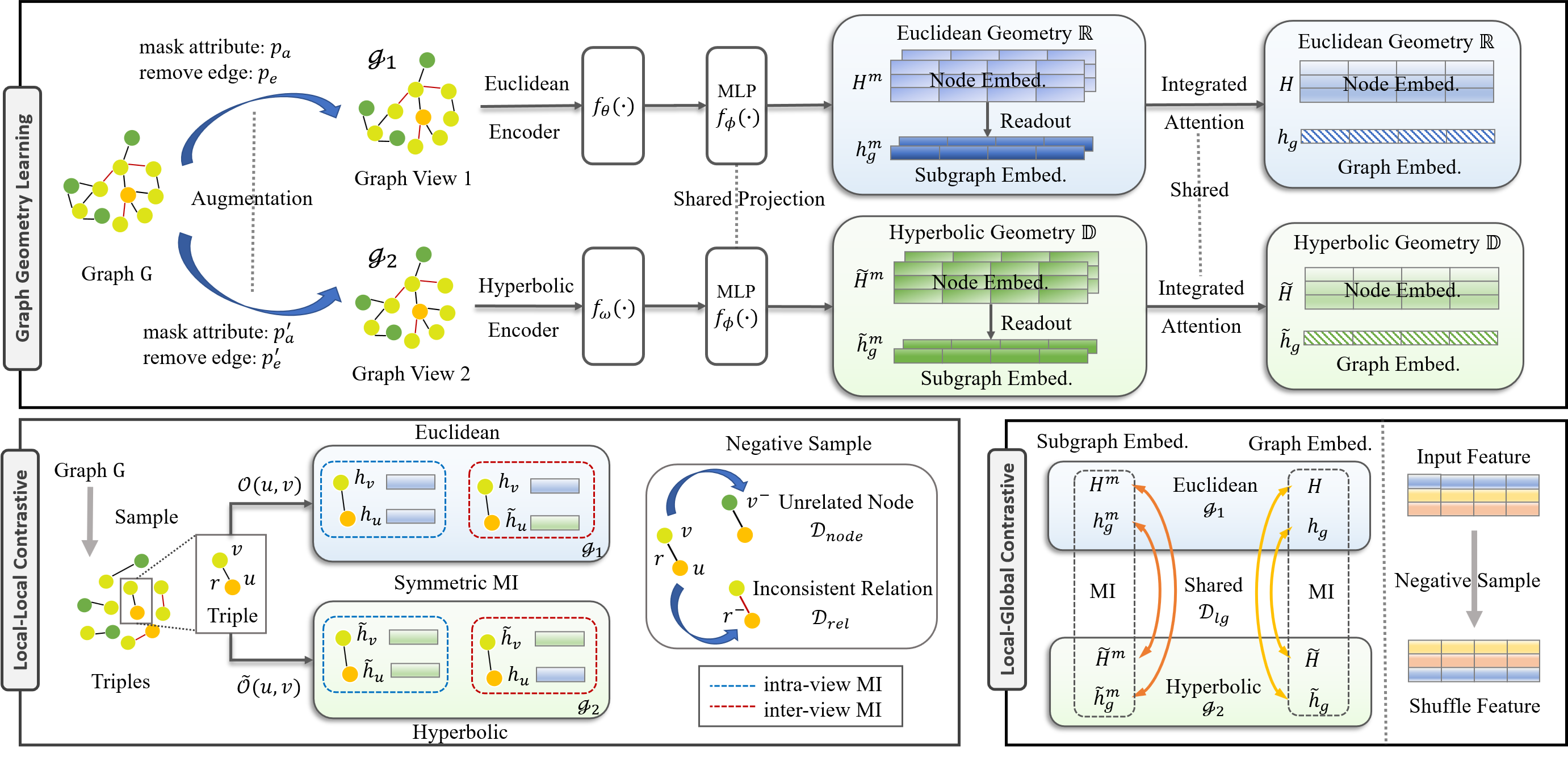}
\caption{Schematic of the GCL architecture. (\expandafter{\romannumeral1}) 
Graph Geometry Learning: generating two graph geometric views, which are embedded in Euclidean space and hyperbolic space, respectively. After generating two graphs $\mathcal{G}_1$ and $\mathcal{G}_2$ via data augmentation, two dedicated graph encoders $f_{\theta}(\cdot)$ and $f_{\omega}(\cdot)$ are applied to embed graphs into different geometric spaces, respectively. The subgraph specific embeddings $\bm{H}^m$ are summed up based on attention mechanism to generate the integrated embeddings $\bm{H}$.
(\expandafter{\romannumeral2}) 
Contrastive Learning Framework: a) Local-Local Contrastive module (see Section \ref{sec:ll-MI}): maximizing MI between one-hop relational connections from inter-view and intra-view, with the aim to discriminate two kinds of negative samples, i.e. unrelated node and inconsistent relation; b) Local-Global Contrastive module (see Section \ref{sec:lg-MI}): maximizing the MI between local and global representations based on meta-path based subgraph $G^m$ and the whole graph $G$.
The method is trained end-to-end.}
\label{fig:model}
\end{figure*}

Contrary to previous works confined in a single geometric space for contrastive learning on heterogeneous graphs, 
in this paper, we define different geometries as specific perspectives under the framework of contrastive learning to provide a more expressive representation for complex structured data.
Geometry Contrastive Learning (GCL) is established as a novel graph contrastive framework to establish relationships between different geometric perspectives, which enforces the geometry invariance in heterogeneous graph neural networks through maximizing agreement between two geometric views of graphs.
Specifically, in our approach, we learn embeddings by maximizing the agreement between different geometric views of heterogeneous graph from both local-local and local-global semantic levels. 
As shown in Figure \ref{fig:model}, our method consists of the following components:
\par
\begin{itemize}
    \item \emph{Graph Geometry Learning}. In order to obtain a more expressive representation of heterogeneous graph, two geometric graph encoders are developed to learn representations from Euclidean and hyperbolic spaces, respectively. The two dedicated encoders operate on different augmented graphs, enriching the diverse contexts of the same heterogeneous graph.
    \item \emph{Contrastive Learning Framework}. The different geometric embeddings learned from graph geometry learning, are then integrated and trained efficiently through the paradigm of contrastive learning. The objective consists of local-local MI and local-global MI.
    1) \emph{Local-Local MI}: maximizing the MI between two views by contrasting node representation pairs under the prior of relation-aware proximity. 2) \emph{Local-Global MI}: maximizing the MI between two views by contrasting local representation from one view and global representation from another view, with respect to the meta-path based subgraphs and the whole graph.
\end{itemize}

\subsection{Graph Geometry Learning}
In graph domain, different geometric spaces provide diverse contexts for each node, especially on heterogeneous graphs.
To further enrich the contexts at both feature and structure levels, we introduce the data augmentation to generate different views, which is 
the key component of contrastive learning methods. 
Thus, in our model, the heterogeneous graph geometry learning consists of two main parts: data augmentation and graph geometric encoder. Specifically, we firstly employ data augmentations to generate two graph views $\mathcal{G}_1$ and $\mathcal{G}_2$. And then two different geometric graph encoders are introduced to give distinctive representations for each view. The different geometric encoders provide inductive biases specific to Euclidean and hyperbolic geometries, which accommodate topologically to heterogeneous graphs very well.

\subsubsection{Data Augmentation}
Following the same spirit of augmentation strategies \cite{you2020graph}, we consider two-level augmentations, including attribute masking for feature-level, and edge permutation for structure-level.
\par
\vspace{2mm}
\noindent\textit{\textbf{Attribute Masking.}}\quad We randomly mask a fraction of dimensions with zeros in node features, to prompt encoders to recover masked nodes attributes using their context information. Specifically, we sample a random vector $\bm{m}\in\mathbb{R}^F$ with the same dimension of initial features, whose entry $m_i$ determines whether to remove the dimension $i$ of node attribute, drawn from Bernoulli distribution. For node attribute $\bm{x}$, the augmented attribute can be computed as
\begin{align}
    \bm{m}_{i} = \text{Bernoulli}(1-p_a),\quad
    \widetilde{\bm{x}} =\bm{x}\circ\bm{m},
\end{align}
where $p_a$ is the probability of each edge being removed, and $\circ$ is Hadamard product.
\par
\vspace{2mm}
\noindent\textit{\textbf{Edge Permutation.}}\quad 
Apart from feature augmentation, we randomly remove a portion of edges in $G$, to construct diverse node contexts for the model to contrast with. Formally, we sample a random masking matrix $S\in\mathbb{R}^{N\times N}$, whose entry $S_{ij}$ determines whether to remove the entry of adjacency $A_{ij}$, drawn from a Bernoulli distribution with probability $p_e$. The resulting adjacency matrix can be computed as
\begin{align}
    S_{ij} = \text{Bernoulli}(1-p_e),\quad
    \widetilde{A} =A\circ S,
\end{align}
where $p_e$ is the probability of each edge being removed, and $\circ$ is Hadamard product.

\subsubsection{Geometric Graph Encoders}
After generating two augmented graphs, they will be fed into two dedicated graph encoders to generate different geometric embeddings located in Euclidean space and hyperbolic space, respectively, in order to capture heterogeneous semantic features and complex structural information from different geometric views.
\par
Benefit from the recent advantages of graph geometry representation learning \cite{chami2019hgcn,liu2019hgnn,zhu2020gil}, our framework allows various choices of the geometric encoders. We opt for simplicity and adopt GCN \cite{kipf2016semi} and HGCN \cite{chami2019hgcn} as our backbone for Euclidean encoder $f_{\theta}(\cdot)$ and hyperbolic encoder $f_{\omega}(\cdot)$, respectively.
The base encoders are designed for each meta-path based subgraph $G^{m}$. The meta-path specific embeddings learned by base encoders are then aggregated based on attention mechanism on account of the importance of each meta-path.
It should be mentioned that all parameters in graph encoders are shared for both positive and negative samples.
\par
\vspace{2mm}
\noindent\textit{\textbf{Euclidean graph encoder.}}\quad For each subgraph $G^{m}$ in graph view $\mathcal{G}_1$, we introduce an Euclidean graph encoder $f_{\theta}:V\rightarrow \mathbb{R}^{d}$ to generate the meta-path specific node embedding $\bm{H}^{m}$. Specifically, we first initialize the node features with input attributes and set the input dimension as $F$. Then, for each layer, the message propagation and update for node $i$ are computed as follows.
\begin{equation}
    \bm{h}_i^{m'}=\sigma\left(\sum_{j\in\{i\}\cup\mathcal{N}_i^{m}} \frac{1}{c_{i,r}} W_{1}^{m} \bm{h}_j^{m}+\bm{b}_1^{m}\right),
\end{equation}
where $W_{1}^{m}\in\mathbb{R}^{d\times d'},\bm{b}_1^{m}\in\mathbb{R}^{d}$ are meta-path $m$ specific trainable matrix and bias in Euclidean graph encoder, where $d', d$ are the dimension of input features and hidden features, respectively. And
$\bm{h}_j^{m}\in\mathbb{R}^{d'}$ denotes the hidden feature vector of neighbor nodes in subgraph $G^{m}$, and $\bm{h}_i^{m'}\in\mathbb{R}^{d}$ denotes the updated hidden features of node $i$ after passing through a message passing layer in Euclidean graph encoder.
The $\mathcal{N}^{m}_i$ denotes the set of neighbor indices of node $i$ in subgraph $G^{m}$, $c_{i,m}$ is a normalization constant that is chosen as $| \mathcal{N}^{m}_i|+1$. 
The global subgraph representation for $G^{m}$ is calculated as the mean of node embeddings with a sigmoid non-linearity, like $\bm{h}_g^{m}=sigmoid\left(\frac{1}{N}\sum_{i=1}^{N}\bm{h}_i^{m}\right)$.
\par
\vspace{2mm}
\noindent\textit{\textbf{Hyperbolic graph encoder.}}\quad 
As mentioned above, the representation power of a single space is limited \cite{sun2015space}, especially for the real-word heterogeneous graphs exhibiting complex non-Euclidean geometries \cite{ravasz2003hierarchical, clauset2008hierarchical}. 
It has been shown that the hyperbolic space offers an efficient alternative to embed non-Euclidean geometry with scale-free or hierarchical structure \cite{ganea2018hyperbolic,chami2019hgcn,zhu2020gil}. 
Therefore, we further introduce the hyperbolic graph encoder that embeds heterogeneous graphs into the hyperbolic space, providing additional inductive biases specific to hyperbolic geometry. 
\par
Specifically, we utilize the Poincar{\'e} ball as the concrete hyperbolic space.
On the augmented graph $\mathcal{G}_2$, 
the hyperbolic graph encoder is defined as $f_\omega: V\rightarrow\mathbb{D}^d_c$, to generate node embeddings $\widetilde{\bm{H}}^{m}\in\mathbb{D}^{N\times d}$ for each subgraph $G^{m}$. And we implement the Poincar{\'e} ball model with $c = 1$.
We first map the features from Euclidean space to hyperbolic space, given that the input features of nodes usually live in Euclidean space. We initialize the node features by performing a \textit{exponential} map $\operatorname{exp}_{\bm{0}}^c(\cdot)$ that maps the input Euclidean feature into hyperbolic space. 
And then we utilize \textit{tangential aggregations} to aggregate neighbor messages in the tangent space via \textit{logarithmic} map $\operatorname{log}_{\bm{0}}^c(\cdot)$, where the mapping functions are defined in HNN \cite{ganea2018hyperbolic}.
\begin{equation}\small
\widetilde{\bm{h}}_i^{m}= \operatorname{exp}_{\bm{0}}^c \left(\sigma\left(\sum_{j\in \{i\}\cup\mathcal{N}_i^{m}} \operatorname{log}_{\bm{0}}^c \left(W_{2}^{m}\otimes_c \widetilde{\bm{h}}_j^{m} \oplus_c \bm{b}_2^{m}\right) \right) \right),
\end{equation}
where $\widetilde{\bm{h}}_i^{m}\in \mathbb{D}_c^{d} \setminus\{\bm{0}\}$ is the hyperbolic hidden features of node $i$ except the zero vector, $W_{2}^{m}\in\mathbb{R}^{d'\times d}$ is a trainable weight matrix, and $\bm{b}_2^{m}\in \mathbb{D}_c^{d}\setminus\{\bm{0}\}$ denotes the hyperbolic bias for subgraph $G^m$. 
For the matrix operation in the Poincar{\'e} ball, 
$\otimes_c$ and $\oplus_c$ denote the corresponding \textit{M$\ddot{o}$bius matrix multiplication} and \textit{M$\ddot{o}$bius addition}, respectively. For the simplicity of notations, we use $M\in\mathbb{R}^{d'\times d}$ denotes the matrix and $\bm{x}, \bm{y}\in \mathbb{D}_c^d \setminus\{\bm{0}\}$ denote the vector to define the \textit{M$\ddot{o}$bius} operations as follows. Please refer to  \cite{ganea2018hyperbolic} for more details.
\begin{align}\small
 M\otimes_c \bm{x} &:= (1/\sqrt{c}) \tanh\left(\frac{\|M\bm{x}\|}{\|\bm{x}\|}\tanh^{-1}(\sqrt{c}\|\bm{x}\|) \right)\frac{M\bm{x}}{\|M\bm{x}\|},\nonumber\\
\bm{x}\oplus_c \bm{y} &:= \frac{(1+2c\langle \bm{x},\bm{y} \rangle+c\lVert \bm{y}\rVert^2)\bm{x}+(1-c\lVert \bm{x}\rVert^2)\bm{y}}{1+2c\langle \bm{x},\bm{y}\rangle + c^2\lVert \bm{x}\rVert^2\lVert \bm{y}\rVert^2}.\nonumber
\end{align}
\par
Besides, the mapping functions used in hyperbolic graph encoder include \textit{exponential} map $\operatorname{exp}_{\bm{x}}^c: \mathcal{T}_{\bm{x}}\mathbb{D}_c^d \rightarrow \mathbb{D}_c^d$ and \textit{logarithmic} map $\operatorname{log}_{\bm{x}}^c: \mathbb{D}_c^d \rightarrow \mathcal{T}_{\bm{x}}\mathbb{D}_c^d$, defined as follows. 
\begin{align}
\operatorname{exp}_{\bm{x}}^c(\bm{v}) &= \bm{x}\oplus_c\left(\operatorname{tanh}\left(\sqrt{c}\frac{\lambda _{\bm{x}}^c \lVert \bm{v}\rVert}{2}\right) \frac{\bm{v}}{\sqrt{c}\lVert \bm{v}\rVert}\right),\nonumber\\
\operatorname{log}_{\bm{x}}^c(\bm{y}) &= \frac{2}{\sqrt{c}\lambda_{\bm{x}}^c}\operatorname{tanh}^{-1}\left( \sqrt{c}\lVert -\bm{x}\oplus_c \bm{y}\rVert \right)\frac{-\bm{x}\oplus_c \bm{y}}{\lVert -\bm{x}\oplus_c \bm{y} \rVert},\nonumber
\end{align}where $\bm{x},\bm{y}\in \mathbb{D}_c^d$, $\bm{v}\in \mathcal{T}_{\bm{x}}\mathbb{D}_c^d$, and $\oplus_c$ denotes the \textit{M$\ddot{\text{o}}$bius addition}.

\par
After aggregating neighbor messages, we apply a pointwise non-linearity $\sigma(\cdot)$, typically ReLU, and then map the embeddings back to hyperbolic space via $\operatorname{exp}_{\bm{0}}^c(\cdot)$. 
For convenience of vector operations between two geometric representations, we further transfer the learned hyperbolic embeddings to vector space via \textit{logarithmic} map, like $\widetilde{\bm{h}}_i^{m}=\operatorname{log}_{\bm{0}}^c(\widetilde{\bm{h}}_i^{m})$. Similarly, we take the mean of node embeddings as hyperbolic graph-level embedding $\widetilde{\bm{h}}_g^{m}$.
\par
After aggregating meta-path specific neighbor information, we will get two node embeddings $\bm{h}^{m}_i$ and $\widetilde{\bm{h}}^{m}_i$ in each subgraph. The embeddings of each view will be summed up to generate the merged embedding $\bm{h}_i$ and $\widetilde{\bm{h}}_i$ based on the important weight $\alpha_i^{m}$ according to the attention mechanism. Here, we utilize the Euclidean embeddings for illustration. 
\begin{equation}\label{eq:attn}
    \bm{h}_i=\sum_{m\in M} \alpha_i^{m}\bm{h}^{m}_i,\quad
     \alpha_i^{m}=\frac{\exp\left(\bm{q}^{m}\cdot\bm{h}_i^{m}\right)}{\sum_{m'\in M}\exp\left(\bm{q}^{m'}\cdot\bm{h}_i^{m'}\right)},
\end{equation}
where $M$ represents the set of meta-paths in $G$, $\alpha_i$ is calculated as the similarity between node feature $\bm{h}_i^{m}$ and a meta-path specific vector $\bm{q}^{m}\in\mathbb{R}^d$. Similarly, the $\widetilde{\bm{h}}_i$, $\bm{h}_g$ and $\widetilde{\bm{h}}_g$ can be integrated in the same manner based on their meta-path specific embeddings. It is worth mentioning that the attention query $\bm{q}^{m}$ is shared among all the merged embeddings.
\par
To sum up, two sets of integrated graph representations are learned by Euclidean and hyperbolic graph encoders, respectively, including node representations $\bm{H}$ and $\widetilde{\bm{H}}$, graph representations $\bm{h}_g$ and $\widetilde{\bm{h}}_g$. 
The learned representations are then fed into a shared projection head $f_{\phi}(\cdot): \mathbb{R}^{N\times d}\rightarrow\mathbb{R}^{N\times d}$, which is a MLP with two hidden layers and an activation. At the inference time, we aggregate the representations learned from two views by averaging them as ultimate embeddings: $\bm{H}_{ult}=\text{mean}[\bm{H},\widetilde{\bm{H}}] \in\mathbb{R}^{N\times d}$.

\subsection{Contrastive Learning Framework}
In this section, we present how to jointly train the two different graph geometric encoders.
The learned embeddings in two geometric spaces are denoted as $\bm{H}=f_{\theta}(X_1,A_1)$ and $\widetilde{\bm{H}}=f_{\omega}(X_2,A_2)$, respectively, where $f_{\theta}(\cdot)$ and $f_{\omega}(\cdot)$ are two dedicate graph encoders. The agreement between different geometric representations is used as the training signal. 
Inspired by recent advances of contrastive learning, we design the geometry contrastive learning framework for heterogeneous graphs that utilizes MI maximization across two incongruent graph geometric views by contrasting learned embeddings at local-local and local-global semantic contexts. 
\par 
\subsubsection{Local-Local MI}\label{sec:ll-MI}
Firstly, we try to make full advantage of the local heterogeneous information by contrastive learning. Different from homogeneous graphs, there exists rich semantics in multiple types of nodes and edges in heterogeneous graphs. Thus, it's imperative to take both topological connectivity and heterogeneous relation into account. Naturally, the local-local MI is formulated to preserve \textit{node consistent} and \textit{relation consistent}.
\par
In contrast to these consistencies, we construct corresponding negative samples from \emph{unrelated nodes} and \emph{inconsistent relations}. Formally, given a positive triple $\langle u,r,v\rangle$ in $G$,  where nodes $u$ and $v$ are connected by relation $r$, we build two sets of negative samples $\mathcal{N}^{node}_{\langle u,v\rangle}$ and $\mathcal{N}^{rel}_{\langle u,r,v\rangle}$ as follows. 
\begin{align}
    &\mathcal{N}^{node}_{\langle u,v\rangle}=\{ \langle u,v^{-} \rangle | (u,v)\in E, (u,v^{-})\notin E, v^{-}\in V \},\label{eq:neg_node} \\
    &\mathcal{N}^{rel}_{\langle u,r,v\rangle}=\{ \langle u,r^{-},v \rangle | (u,v)\in E, \phi_e(u,v)=r, r^{-}\in R\setminus r \}.\label{eq:neg_rel}
\end{align}
The node pair $\langle u,v^{-}\rangle$ in Eq.~(\ref{eq:neg_node}), denoting that there are no edges between $u$ and $v^{-}$, is to serve as the negative samples of adjacent nodes.
And the triple $\langle u,r^{-},v \rangle$ in Eq.~(\ref{eq:neg_rel}) indicates that there is no relation $r^{-}$ between nodes $u$ and $v$, which represents the different semantic context from $\langle u,r,v\rangle$.
\par
Different from the negative samples constructed in recent works, such as PT-HGNN \cite{jiang2021pthgnn}, which employs node instances and network schema instances as self-supervision, we construct heterogeneous node instances $\mathcal{N}^{node}_{\langle u,v\rangle}$ and relation instances $\mathcal{N}^{rel}_{\langle u,r,v\rangle}$ as two types of negative samples. 
For the above two kinds of negative samples, we design the corresponding discriminator with different contrastive objective via maximizing the MI between positive samples. That is \textit{Node Consistent} and \textit{Relation Consistent} on account of negative samples $\mathcal{N}^{node}_{\langle u,v\rangle}$, and $\mathcal{N}^{rel}_{\langle u,r,v\rangle}$, respectively.
\par
\vspace{2mm}
\noindent\textit{\textbf{Node Consistent.}} 
To ensure the consistency between connected nodes $u$ and $v$, previous studies maximized MI between latent representations of connected nodes.
To enrich the local information in heterogeneous graphs, a recent work HeCo \cite{wang2021cocontrastive} has extended the single view contrastive learning into dual-views. Different from the work that maximizes local mutual information from inter-view, we extend the dual-view contrast into both inter-view and intra-view across two geometric graph views.
Specifically, given the node pair $\langle u,v\rangle$, the agreement of two node embeddings in the same geometric space can be seen as intra-view MI, while the agreement of embeddings from different geometric spaces is regarded as inter-view MI, formally given by,
\begin{equation}\label{eq:MI_node_1}
    \mathscr{O}\left(u,v\right)=
    \underbrace{MI(\bm{h}_u, \bm{h}_v)}_{\text{intra-view MI}} + \underbrace{MI(\bm{h}_u, \widetilde{\bm{h}}_v)}_{\text{inter-view MI}},
\end{equation}
where $\bm{h}_u$ and $\bm{h}_v$ are node embeddings generated in one view, and $\widetilde{\bm{h}}_v$ is generated in another view. Since two geometric views are symmetric, the MI of another view can be defined with,
\begin{equation}\label{eq:MI_node_2}
    \widetilde{\mathscr{O}}\left(u,v\right)=
    \underbrace{MI(\widetilde{\bm{h}}_u, \widetilde{\bm{h}}_v)}_{\text{intra-view MI}} + \underbrace{MI(\widetilde{\bm{h}}_u, \bm{h}_v)}_{\text{inter-view MI}}.
\end{equation}
The objective of node-level agreement $\mathcal{O}_{node}$ to be maximized is then defined as the average over all positive pairs as follows.
\begin{equation}\label{eq:loss_node}
      \mathcal{O}_{node} = \frac{1}{2|\mathcal{P}^{node}|}\sum_{\langle u,v\rangle\in \mathcal{P}^{node}} \left[\mathscr{O}\left(u,v\right)+\widetilde{\mathscr{O}}\left(u,v\right) \right],
\end{equation}
where $\mathcal{P}^{node}$ is the set of positive node pairs sampled from graph $G$, and $|\mathcal{P}^{node}|$ is the number of positive samples.
\par
As a proxy for maximizing the consistency, we employ a discriminator $\mathcal{D}_{node}:\mathbb{R}^{d}\times\mathbb{R}^{d}\rightarrow\mathbb{R}$, to model the agreement probability scores assigned to node representation pairs from intra-view and inter-view. 
Specifically, we implement the discriminator as a bilinear function. Here we utilize the intra-view embeddings for illustration.
\begin{equation}
    \mathcal{D}_{node}(\bm{h}_u, \bm{h}_v)=\sigma\left( \bm{h}_u^{\top} W_{node} \bm{h}_v \right),
\end{equation}
where $W_{node}\in \mathbb{R}^{d\times d}$ is a trainable weight matrix, and $\sigma$ is a sigmoid non-linearity.
\par
\vspace{2mm}
\noindent\textit{\textbf{Relation Consistent.}} When it comes to relation consistent, the MI of representations are also considered as objective in Eq.~(\ref{eq:MI_rel}). It is imperative to consider heterogeneity existing in relational connectivity. Different from node consistent, the relations are introduced to estimate the MI between triple $\langle u,r,v\rangle$.
\begin{equation}\label{eq:MI_rel}
    \begin{split}
    \mathscr{O}\left(u,r,v\right)&=
    \underbrace{MI(\bm{h}_u, r, \bm{h}_v)}_{\text{intra-view MI}} + \underbrace{MI(\bm{h}_u, r, \widetilde{\bm{h}}_v)}_{\text{inter-view MI}},\\
    \widetilde{\mathscr{O}}\left(u,r,v\right)&=
    \underbrace{MI(\widetilde{\bm{h}}_u, r, \widetilde{\bm{h}}_v)}_{\text{intra-view MI}} + \underbrace{MI(\widetilde{\bm{h}}_u, r, \bm{h}_v)}_{\text{inter-view MI}}.
    \end{split}
\end{equation}
Therefore, the objective for relation consistent objective $\mathcal{O}_{rel}$ to be maximized is defined based on the symmetric MI as follows.
\begin{equation}\label{eq:loss_rel}
      \mathcal{O}_{rel}= \frac{1}{2|\mathcal{P}^{rel}|}\sum_{\langle u,r,v\rangle\in \mathcal{P}^{rel}} \left[\mathscr{O}\left(u,r,v\right)+\widetilde{\mathscr{O}}\left(u,r,v\right) \right],
\end{equation}
where $|\mathcal{P}^{rel}|$ is the number of sampled positive triples $\langle u, r, v\rangle$ in graph $G$. 
The discriminator function is designed as $\mathcal{D}_{rel}:\mathbb{R}^{d}\times\mathbb{R}\times\mathbb{R}^{d}\rightarrow\mathbb{R}$ to model MI between triple $\langle u,r,v\rangle$, which scores the agreement between the relation-aware node pairs in two views. We implement the discriminator as the relation-aware bilinear function as follows, using intra-view embeddings for illustration
\begin{align}
    \mathcal{D}_{rel}(\bm{h}_u, r, \bm{h}_v)=\sigma\left( \bm{h}_u^{\top} W_{rel}^{(r)} \bm{h}_v \right),
\end{align}where $r$ denotes the relation type connecting node $u$ and $v$, computed as $\phi_e(u,v)$. $W_{rel}^{(r)}\in \mathbb{R}^{d\times d}$ is a trainable relation-specific weight matrix, and $\sigma$ is a sigmoid non-linearity. 
\par
\subsubsection{Local-Global MI}\label{sec:lg-MI}
Besides the mutual information between local node representations, it is important to take the agreement between node and graph representations into account.
This encourages the encoder to carry the type of heterogeneous information that appear globally relevant, such as the case of a class label.
In order to capture the global heterogeneous knowledge comprehensively, we extend the high-level global representation to two parts, including the meta-path based subgraphs $\bm{h}_{g}^{m}$ and the whole graph $\bm{h}_{g}$. 
We empirically show that each part has a positive effect on the representation of heterogeneous graphs (see section \ref{sec:exp_analysis}).
The objective $\mathcal{O}_{lg}$ to be maximized for local-global MI is then defined as follows.
\begin{equation}\label{eq:loss_lg}
    \mathcal{O}_{lg}=
    \frac{1}{N}\sum_{i=1}^{N} 
    \left[
    \frac{1}{|M|}\sum_{m\in M}\left[ MI(\bm{h}_i^{m}, \widetilde{\bm{h}}_g^{m}) + MI(\widetilde{\bm{h}}_i^{m}, \bm{h}_g^{m}) \right]
    + \left[ MI(\bm{h}_i, \widetilde{\bm{h}}_g) + MI(\widetilde{\bm{h}}_i, \bm{h}_g) \right] 
    \right],
\end{equation}
where $N$ and $|M|$ are the number of nodes and meta-paths respectively. $\bm{h}^{m}_i$ and $\bm{h}^{m}_g$ are meta-path specific embeddings for node and graph. $\bm{h}_i$ and $\bm{h}_g$ are integrated embeddings. In addition to MI between integrated embeddings at the whole graph level, subgraph specific local-global MIs are also introduced to capture the meta-path based global information in heterogeneous graphs.
\par
It is worth noting that the difference between our method and related works (i.e. MVGRL \cite{hassani2020contrastive}, HDGI \cite{ren2019heterogeneous} and DMGI \cite{park2020unsupervised}) lies in local-global MIs. These methods only take one part of the local-global InfoMax objective into account. HDGI considers the subgraph specific local-global MIs, while MVGRL and DMGI maximize the MI by contrasting node representations with graph representation in a whole graph. The detailed model comparison can be seen in the section \ref{sec:comp}.
\par
Similarly, we utilize a discriminator $\mathcal{D}_{lg}:\mathbb{R}^{d}\times\mathbb{R}^{d}\rightarrow\mathbb{R}$ to model MI, which takes in a node representation from one view and a graph representation from another view. We implement it as a bilinear function with a trainable weight matrix $W_{lg}\in\mathbb{R}^{d\times d}$ in Eq.~(\ref{eq:d_lg}). It is worth mentioning that $W_{lg}$ is shared among all local-global MIs in Eq.~(\ref{eq:loss_lg}), enabling to learn a universal discriminator that can maximize the MI between positive pairs regardless global levels.
\begin{equation}\label{eq:d_lg}
    \mathcal{D}_{lg}(\bm{h}_i,\widetilde{\bm{h}}_g)=\sigma\left(\bm{h}_i^{\top} W_{lg} \widetilde{\bm{h}}_g\right).
\end{equation}
\subsubsection{Contrastive Regularization}\label{sec:reg}
Considering that the multiple subgraphs in hetereogeneous graphs contain different semantic knowledge, it is necessary to facilitate the agreement of node embeddings regarding subgraphs by comparing negative samples. Different from HDGI \cite{park2020unsupervised} that introduces an extra consensus embedding matrix, we utilize the ultimate merged node embedding $\bm{H}_{ult}$ as consensus embedding, to minimize the disagreement among the meta-path specific node embeddings $\bm{H}^{m}_{ult}$ and $\bm{H}_{ult}$.
Specifically, we introduce a regularization to not only maximize the similarity between positive node embeddings $\bm{H}^{m}_{ult}$ and the merged node embedding $\bm{H}_{ult}$, but also minimize the similarity between negative node embeddings $\bm{H}^{m}_{neg}$ and $\bm{H}_{ult}$. Formally,
\begin{equation}\label{eq:loss_reg}
    \mathcal{L}_{reg}=\sum_{m\in M}\left[\left( \bm{H}_{ult}-\bm{H}^{m}_{ult} \right)^{2}-\left( \bm{H}_{ult}-\bm{H}_{neg}^{m} \right)^{2}\right].
\end{equation} 
Here we take the same sampling strategy for $\bm{H}_{neg}^{m}$ as the negative sample generator in Section \ref{sec:neg_sample}. 
\subsection{Model Training}
\subsubsection{Estimator}
The local-local MI and local-glocal MI are uniformly estimated by binary cross-entropy loss, which is an effective estimator based on Jensen-Shannon divergence between the joint distribution and the product of marginals. It is formulated based on close connections between generative adversarial network (GAN) and Jensen-Shannon divergence \cite{Nowozin2016fgan}.
The concrete loss function to be minimized is defined as follows.
\begin{equation}
    \mathcal{L}_{*} = -\frac{1}{|pos|+|neg|}
    \left( 
    \sum \mathbb{E}_{pos} \left[\log \mathcal{D}_{*} \right] +
    \sum \mathbb{E}_{neg} \left[\log \left( 1 - \mathcal{D}_{*} \right)\right]\right),
\end{equation}
where $pos$ and $neg$ denote the positive and negative sets, and the number of each set is denoted as $|\cdot|$. $\mathcal{D}_{*}$ is the corresponding discriminator to score the agreement between embeddings, referring to $\mathcal{D}_{node}$, $\mathcal{D}_{rel}$ and $\mathcal{D}_{lg}$. Therefore, we can obtain the concrete loss for each MI objective defined in Eq.~(\ref{eq:loss_lg}), Eq.~(\ref{eq:loss_node}), and Eq.~(\ref{eq:loss_rel}), denoted as $\mathcal{L}_{lg}$, $\mathcal{L}_{node}$ and $\mathcal{L}_{rel}$, respectively.
\par
\subsubsection{Negative sample generator}\label{sec:neg_sample}
The positive samples are generated from joint distribution 
and the negative samples are generated from the product of marginals.
In local-local contrastive learning, at each epoch, we randomly sample the fixed size of positive triples. For negative samples in discriminators $\mathcal{D}_{node}$ and $\mathcal{D}_{rel}$, we sample the same number of triples, based on the guidance of negative sampling strategies $\mathcal{N}^{node}_{\langle u,v\rangle}$ and $\mathcal{N}^{rel}_{\langle u,r,v\rangle}$ defined in Eq.~(\ref{eq:neg_node}) and Eq.~(\ref{eq:neg_rel}), respectively.
In local-global contrastive learning, we randomly shuffle the node features as the negative node embeddings, following the common settings in \cite{velickovic2019deep}. 
\par
To sum up, we jointly optimize the sum of all concrete losses. The overall loss $\mathcal{L}$ to be minimized is computed as follows.
\begin{equation}\label{eq:loss}
    \mathcal{L}=\mathcal{L}_{lg}+\beta\mathcal{L}_{node}+\lambda\mathcal{L}_{rel} + 
    \gamma\mathcal{L}_{reg},
\end{equation}
where $\beta$, $\lambda$ and $\gamma$ are hyperparameters to control the contribution of each part, and $\mathcal{L}_{reg}$ is the contrastive regularization term (see Section \ref{sec:reg}). The learning algorithm is summarized in Algorithm \ref{algorithm}.

\subsection{Model Analysis}\label{sec:model_analysis}
\subsubsection{Time Complexity}
The main operations of our model are learning node representations and calculating the total loss. To avoid parameter overhead, we use the shared projection head, attention query and discriminator in the model.
For graph encoder module, the time complexity is $O(NFd+|E|d)$, where $N$ and $|E|$ are the numbers of nodes and edges, $F$ and $d$ are the dimension of input and hidden features, respectively. 
\begin{algorithm}[t!]
 \caption{Geometry Contrastive Learning on HG Algorithm}
 \label{algorithm}
 \begin{algorithmic}[1]
  \REQUIRE ~~\\ Graph $G=(V,E,T,R)$, initial features $X$, hidden dimension $d$, graph encoders $f_{\theta}(\cdot)$, $f_{\omega}(\cdot)$, projection head $f_{\phi}(\cdot)$, sample size $s$.
  \ENSURE ~~\\
  The trained model, i.e., the encoder parameters $\theta$ and $\omega$, $\phi$ and ultimate learned embeddings $\bm{H}_{ult}$.\\
  \STATE {Randomly initialize the model parameters $\theta$, $\omega$, and $\phi$}.
  \FOR {$i\in [1,num\_of\_epoches]$}
    \STATE {Generate two graph views $\mathcal{G}_1$ and $\mathcal{G}_2$ by performing augmentations on $G$}.
    \STATE {Obtain meta-path specific embeddings $\bm{H}^{m}$, $\bm{h}_g^{m}$, and the merged embeddings $\bm{H}$, $\bm{h}_g$ of $\mathcal{G}_1$ using encoder $f_{\theta}(\cdot)$}.
    \STATE {Obtain meta-path specific embeddings $\widetilde{\bm{H}}^{m}$, $\widetilde{\bm{h}}_g^{m}$ and the merged embeddings $\widetilde{\bm{H}}$, $\widetilde{\bm{h}}_g$ of $\mathcal{G}_2$ using encoder $f_{\omega}(\cdot)$}.
    \STATE {Project two geometric node embeddings}.
    \STATE {Compute local-global contrastive objective $\mathcal{L}_{lg}$}.
    \COMMENT{Eq.~(\ref{eq:loss_lg})}
    \STATE{Sample $s$ positive triples in graph $G$}.
        \FOR{\textit{each positive triple} $\langle u,r,v\rangle$}
        \STATE {Sample from $\{v^{-}|\langle u,v^{-} \rangle\in\mathcal{N}^{node}_{\langle u,v\rangle}\}$ and compute node contrastive loss}. \COMMENT{Eq.~(\ref{eq:loss_node})}
        \STATE {Sample from $\{r^{-}|\langle u,r^{-},v \rangle\in\mathcal{N}^{rel}_{\langle u,r^{-},v\rangle}\}$ and compute relation contrastive loss}. \COMMENT{Eq.~(\ref{eq:loss_rel})}
        \ENDFOR
    \STATE {Calculate the average of learned embeddings from two views as ultimate embeddings $\bm{H}_{ult}$, $\bm{H}_{ult}^{m}$.}
    \STATE {Compute regularization loss $\mathcal{L}_{reg}$}.
    \COMMENT{Eq.~(\ref{eq:loss_reg})}
    \STATE {Compute the overall losses $\mathcal{L}$}.
    \COMMENT{Eq.~(\ref{eq:loss})}
    \STATE {Update model parameters using stochastic gradient descent}.
    \IF {convergence}
        \STATE {break loop}.
    \ENDIF
  \ENDFOR
 \end{algorithmic}
\end{algorithm}

\subsubsection{Theoretical Analysis}\label{sec:disc}
The designed method is to recognize rich semantics and geometric structural properties in heterogeneous graphs by introducing a comprehensive geometric views and training with the local-global InfoMax objective function.
Intrinsically, after obtaining dual-view graph geometric representations at both local and global levels, our InfoMax objective can be reformulated as noise-contrastive losses, where positive samples come from the joint distribution $x_p\sim p([\bm{H}, \mathcal{G}_1],[\widetilde{\bm{H}}, \mathcal{G}_2])$ and negative samples come from the product of marginals $x_n\sim p\left([\bm{H}, \mathcal{G}_1]\right)p([\widetilde{\bm{H}}, \mathcal{G}_2])$.
For local-local contrastive objective, we provide some intuition that connects our local-local loss and MI maximization between the input node features and the learned node embeddings in two geometric spaces. Formally, given two random variables $\bm{h}_i,\widetilde{\bm{h}}_i$ as the embedding in the two views, with their joint distribution denoted as $p([\bm{h}_i, \mathcal{G}_1], [\widetilde{\bm{h}}_i, \mathcal{G}_2])$, our local-local objective is a lower bound of MI between encoder input $\bm{x}_i$ and node representations in two geometric embeddings $\bm{h}_i,\widetilde{\bm{h}}_i$. 
For local-global contrastive objective, it effectively maximizes MI between node embeddings and global embeddings, based on the Jensen-Shannon divergence between the joint distribution $x_p\sim p\left([\bm{h}_i, \mathcal{G}_1],[\widetilde{\bm{h}}_g, \mathcal{G}_2]\right)$ and the product of marginals $x_n\sim p([\bm{h}_i, \mathcal{G}_1])p([\widetilde{\bm{h}}_g, \mathcal{G}_2])$.

\subsubsection{The Connection with Multi-view Learning}
Intrinsically, the designed method is to recognize complex structures and semantics existing in graph heterogeneity.
As stated above, the two prominent characteristics in heterogeneous graphs, rich semantics and complex structures, make the representation learning on a single space more challenging. 
The proposed method GCL first introduces two different geometries as the feature spaces, providing an alternative mixed-curvature space for complex heterogeneous graphs.
The Euclidean and hyperbolic geometries can be regarded as two different graph views that are inclined to capture regular and hierarchical semantics, respectively. It takes the advantages of both powerful simplicity and efficiency of Euclidean geometry and complex network representational capability of hyperbolic geometry.
Besides, under the framework of contrastive learning, the two graph geometric embeddings can be further described as different views.
The connection with multi-view learning defined above provides the same context and convenience for the model comparisons below.

\subsubsection{Model Comparison}\label{sec:comp}
In the model comparison, we mainly focus on the difference of the InfoMax objective and dual-view design. In Table \ref{tab:baselines}, we provide a summary of the model capability of our proposed method GCL with other related methods. 
There are some important design differences between them. 
1) InfoMax objective is different. GCL focuses on not only local-local InfoMax, but also local-global InfoMax. Besides, the local-global InfoMax objective is further extended to two global levels, the whole graph and meta-path based subgraph. The multi-level graph representations are capable of effectively preserving rich interactions of heterogeneous graphs. By contrast, other methods are designed only from one aspect, either local-local or local-global.
2) Dual-view design is different. Considering the complex structural characteristics of heterogeneous graphs, we introduce the different graph geometries as dual-view to provide different diverse geometric perspectives across incongruent graphs. 
While other works are confined in a single space, and they focus primarily on data augmentation design for multiple perspectives.
Overall, our method is the most comprehensive one which encode heterogeneous self-supervised signals from multiple levels.
\begin{table*}[htbp]
\caption{Comparison of our proposed method with other related methods. For each method, we depict whether the method can model heterogeneous graphs, the components of the objective function, and whether it contains multiple graph views.}
\centering\label{tab:baselines}
\resizebox{.99\textwidth}{!}{
\begin{tabular}{lcccccc}
\toprule
Methods & Heterogeneous & Unsupervised & Local-Local MI & Local-Global MI & sub-Local-Global MI & Multi-view \\
\midrule
  DGI \cite{velickovic2019deep} & \xmark & \checkmark & \xmark & \checkmark & \xmark & \xmark \\
  GRACE \cite{zhu2020deep} & \xmark & \checkmark & \checkmark & \xmark & \xmark & \xmark \\
  MVGRL \cite{hassani2020contrastive} & \xmark & \checkmark & \xmark & \checkmark & \xmark & \checkmark \\
  HDGI \cite{ren2019heterogeneous} & \checkmark & \checkmark & \xmark & \xmark & \checkmark & \xmark \\
  DMGI \cite{park2020unsupervised} & \checkmark & \checkmark & \xmark & \checkmark & \xmark & \xmark \\
  PT-HGNN \cite{jiang2021pthgnn} & \checkmark & \checkmark & \checkmark & \xmark & \xmark & \xmark \\
  HeCo \cite{wang2021cocontrastive} & \checkmark & \checkmark & \checkmark & \xmark & \xmark & \checkmark \\
\midrule
  Our method & \checkmark & \checkmark & \checkmark & \checkmark & \checkmark & \checkmark \\
\bottomrule
\end{tabular}
}
\end{table*}

\section{Experiments}\label{sec:exp}
In this section, we conduct extensive experiments to verify the performance of our model on node classification, node clustering and similarity search tasks, compared with both state-of-the-art unsupervised and supervised methods. Code and data are available at \url{https://github.com/hete-graph/CMHG}.
\subsection{Experiments Setup}
\subsubsection{Datasets} 
We consider four benchmark data sets: DBLP, IMDB, ACM and Amazon. Dataset statistics and details are summarized in Table \ref{tab:datasets}. DBLP \cite{park2020unsupervised} is a research paper set collected from computer science bibliographies, where the paper can be classified into four classes: DM, AI, CV, NLP. The meta-paths defined in DBLP include Paper-Author-Paper (PAP), Paper-Paper-Paper (PPP) and Paper-Author-Term-Author-Paper (PATAP).
IMDB \cite{park2020unsupervised} is a knowledge graph of movies containing three types of movies. The meta-paths are Movie-Actor-Movie (MAM), Movie-Director-Movie (MDM) and Movie-Keyword-Movie (MKM).
ACM \cite{wang2019heterogeneous} is another academic paper dataset in which the meta-paths are defined as: Paper-Author-Paper (PAP) and Paper-Subject-Paper (PSP). 
Amazon \cite{park2020unsupervised} is a dataset that contains a multiplex network of items, i.e., also-viewed (IVI), also-bought (IBI) and bought-together (ITI). The task is to classify the items into four classes: Beauty, Automotive, Patio Lawn and Garden, and Baby.
Table \ref{tab:datasets} shows the statistics of four benchmarks used in the paper, where nodes have attributes using bag-of-words of text associated with each node. 
We further compute the hyperbolicity $\delta$ of the maximum connected subgraph for each meta-path based graph to characterize the degree of how tree-like a graph exists over the graph, defined by \cite{gromov1987hyperbolic}. And the lower hyperbolicity is, the more hyperbolic the graph is. For example, a tree-like graph is with 0-hyperbolic.
As shown in Table \ref{tab:datasets}, the datasets exhibit multiple different hyperbolicities in a graph, which are suitable for capturing effective information from multiple geometric perspectives.
\begin{table*}[htbp]
\caption{Datasets overview. For each dataset, we depict the nodes types, meta-paths, and corresponding numbers. In details, we depict the dimension of nodes attributes, number of classes and the dataset split in Train/Validation/Test.}
\centering
\resizebox{.99\textwidth}{!}{
\begin{tabular}{c|c|ccc|c|c|c}
\toprule
Datasets & Nodes &  Meta-path & Num. & Hyperbolicity $\delta$ & Num. node attributes & Num. classes & Train/Validation/Test\\
\midrule
  DBLP & \makecell[l]{Paper (P): 14,328 \\ Author (A): 9,867 \\ Term (T): 1,975} & \makecell[c]{P-A-P\\P-P-P\\P-A-T-A-P} & \makecell[c]{144,783\\90,145\\57,137,515} & \makecell[c]{4\\2.5\\1} & 2,000 & 4 & 80/200/7,627 \\
\midrule
  IMDB & \makecell[l]{Movie (M): 3,550 \\ Actor (A): 4,441 \\ Director (D): 1,726} & \makecell[c]{M-A-M\\M-D-M} & \makecell[c]{66,428\\13,788} & \makecell[c]{3\\0} & 1,007 & 3 & 300/300/2,950 \\
\midrule
  ACM & \makecell[l]{Paper (P): 3,025 \\ Author (A): 5,835 \\ Subject (S): 56} & \makecell[c]{P-A-P\\P-S-P} & \makecell[c]{29,281\\2,210,761} & \makecell[c]{4\\0} & 1,830 & 3 & 600/300/2,125 \\
\midrule
  Amazon & Item (I): 7,621 & \makecell[c]{I-V-I\\I-B-I\\I-T-I} & \makecell[c]{266,237\\1,104,257\\16,305} & \makecell[c]{3\\2.5\\1.5} & 2,000 & 4 & 80/200/7,341 \\
\bottomrule
\end{tabular}
}
\label{tab:datasets}
\end{table*}

\subsubsection{Baselines}
We compare our method with several state-of-the-art baselines, including 8 unsupervised methods and 5 supervised methods, to evaluate the effectiveness of our method. The state-of-the-art methods can be further divided into two categories according to whether the method is designed for homogeneous graph or heterogeneous graph. The details are as follows. To give a comprehensive comparison of baselines, we provide a summary of the properties of the compared methods in Table \ref{tab:baselines}.
\par
\begin{itemize}
    \item Unsupervised methods: 
    \par
    (1) \textbf{Homogeneous graph.} Baselines designed for homogeneous graphs on the unsupervised manner contain shallow methods including Raw Feature, Deepwalk \cite{perozzi2014deepwalk} and Node2vec \cite{grover2016node2vec}, and deep methods including DGI \cite{velickovic2019deep}, GRACE \cite{zhu2020deep} and MVGRL \cite{hassani2020contrastive}. The shallow methods are mainly network embedding models based on random walk and skip-gram, and Raw Feature uses the initial features as embeddings. DGI \cite{velickovic2019deep} maximizes MI between node and graph representations for learning a better graph encoder. GRACE proposes a hybrid scheme for generating graph views on both structure and attribute levels. MVGRL \cite{hassani2020contrastive} maximizes MI between representations encoded from different structural views of graphs. 
    \par
    (2) \textbf{Heterogeneous graph.} For heterogeneous baselines, we compare our method with two strong unsupervised methods: HDGI \cite{ren2019heterogeneous} and DMGI \cite{park2020unsupervised}. HDGI \cite{ren2019heterogeneous} maximizes the local-global MI for learning a heterogeneous graph encoder. DMGI \cite{park2020unsupervised} jointly integrates the MI objective and relation specific node embeddings through a consensus regularization to train the graph encoder. Note that, as we only consider the meta-path specific adjacency matrix, the network schema based methods are not compared with here such as PT-HGNN \cite{jiang2021pthgnn} and HeCo \cite{wang2021cocontrastive}, which require the additional knowledge of multi-type nodes, i.e., Paper, Author, Term. Here, we only consider one type node, i.e., Paper. 
    
    \item Supervised methods: 
    \par
    (1) \textbf{Homogeneous graph.} GCN \cite{kipf2016semi} and GAT \cite{velivckovic2017graph} are two representative supervised GNN-based methods designed for homogeneous graphs. GCN \cite{kipf2016semi} is a semi-supervised methods for node classification on homogeneous graphs. Based on GCN, GAT \cite{velivckovic2017graph} applies the attention mechanism in homogeneous graphs for node classification. 
    \par
    (2) \textbf{Heterogeneous graph.} We compare our method with three strong supervised methods including R-GCN \cite{schlichtkrull2018modeling}, mGCN \cite{ma2019multi} and HAN \cite{wang2019han}. R-GCN considers different relations by learning multiple convolutional matrices. mGCN applies GCN and GAT on multiplex network to embed inter-network and intra-network interactions. HAN utilizes meta-paths to model higher-order proximity and employs node-level and semantic-level attentions. Similarly, the relevant heterogeneous methods are not compared here which are designed for multi-type nodes, such as HGT \cite{hu2020heterogeneous}.
\end{itemize}

\par

\begin{table*}[htbp]
\centering
\caption{Node classification results for supervised and unsupervised models in Macro-F1 and Micro-F1 on four datasets (\%). The training data column highlights the data available to each model during training ($X$: features, $A$: adjacency matrix, $Y$: labels).}
\resizebox{.99\textwidth}{!}{
\label{tab:nc_results}
\begin{tabular}{llccccccccccccc}
\toprule
& \multirow{2}{*}{Method} & \multirow{2}{*}{Training Data} & \multicolumn{2}{c}{ACM} &
\multicolumn{2}{c}{IMDB} &
\multicolumn{2}{c}{DBLP} & 
\multicolumn{2}{c}{Amazon} \\
\cmidrule(lr){4-5}\cmidrule(lr){6-7}\cmidrule(lr){8-9}\cmidrule(lr){10-11} & & & MacF1 & MicF1 & MacF1 & MicF1 & MacF1 & MicF1 & MacF1 & MicF1 \\ 
\midrule
\multirow{5}{*}{\rotatebox{90}{SUPERVISED}}
  & GCN & $X$, $A$, $Y$ & 84.74$\pm$0.63 & 84.82$\pm$0.56 & 61.43$\pm$0.57 & 61.99$\pm$0.54 & 69.44$\pm$0.84 & 68.64$\pm$0.50 & 63.60$\pm$0.95 & 63.97$\pm$0.85 \\
  & GAT & $X$, $A$, $Y$ & 85.41$\pm$0.59 & 85.50$\pm$0.54 & 60.38$\pm$0.49 & 61.12$\pm$0.46 & 72.48$\pm$0.60 & 71.35$\pm$0.51 & 51.93$\pm$0.36 & 52.71$\pm$0.36 \\
  & R-GCN & $X$, $A$, $Y$ & 86.01$\pm$0.26 & 86.08$\pm$0.25 & 61.53$\pm$0.73 & 62.14$\pm$0.66 & 69.43$\pm$1.96 & 69.01$\pm$1.48 & 64.01$\pm$1.20 & 64.51$\pm$1.07\\
  & mGCN & $X$, $A$, $Y$ & 85.80$\pm$0.43 & 86.06$\pm$0.45 & 62.34$\pm$0.68 & 63.08$\pm$0.38 & 72.16$\pm$1.02 & 71.29$\pm$0.93 & 65.92$\pm$0.46 & 66.10$\pm$0.77 \\
  & HAN & $X$, $A$, $Y$ & 87.83$\pm$0.62 & 89.24$\pm$0.59 & 60.04$\pm$0.50 & 60.30$\pm$0.47 & 71.61$\pm$1.23 & 70.75$\pm$0.89 & 53.43$\pm$0.80 & 54.45$\pm$0.75 \\
\midrule
\midrule
  \multirow{9}{*}{\rotatebox{90}{UNSUPERVISED}}
  & Raw features & $X$ & 71.00$\pm$0.55 & 72.13$\pm$0.45 & 56.44$\pm$0.90 & 57.06$\pm$0.70 & 51.88$\pm$0.54 & 51.70$\pm$0.69 & 70.30$\pm$0.75 & 70.58$\pm$0.83 \\
  & Deepwalk & $A$ & 73.91$\pm$0.69 & 74.89$\pm$0.53 & 52.33$\pm$0.67 & 54.60$\pm$0.52 & 53.34$\pm$0.67 & 53.67$\pm$0.70 & 66.31$\pm$0.79 & 67.09$\pm$0.81 \\
  & Node2vec & $A$ & 74.12$\pm$0.63 & 74.88$\pm$0.56 & 53.40$\pm$0.72 & 54.76$\pm$0.63 & 54.31$\pm$0.72 & 54.69$\pm$0.81 & 66.21$\pm$0.77 & 67.01$\pm$0.75 \\
  & DGI & $X$, $A$ & 87.47$\pm$1.68 & 87.73$\pm$1.70 & 57.05$\pm$0.78 & 58.39$\pm$0.59 & 74.28$\pm$1.22 & 73.78$\pm$1.51 & 40.28$\pm$1.41 & 40.37$\pm$1.33 \\
  & GRACE & $X$, $A$ & 84.28$\pm$1.43 & 84.39$\pm$1.47& 63.48$\pm$0.89 & 64.07$\pm$0.80 & 72.40$\pm$1.27 & 71.13$\pm$1.32 & 67.07$\pm$1.39 & 67.64$\pm$1.45 \\
  & MVGRL & $X$, $A$ & 87.90$\pm$1.45 & 88.03$\pm$1.38 & 58.31$\pm$1.12 & 59.30$\pm$0.93 & 74.81$\pm$1.63 & 73.08$\pm$1.96 & 38.38$\pm$1.35 & 38.92$\pm$1.27 \\
  & HDGI & $X$, $A$ & 89.31$\pm$1.56 & 89.30$\pm$1.42 & 59.47$\pm$1.19 & 60.83$\pm$0.89 & 78.41$\pm$0.79 & 77.95$\pm$0.82 & 42.93$\pm$1.84 & 43.73$\pm$1.97\\
  & DMGI & $X$, $A$ & 87.77$\pm$1.35 & 88.00$\pm$1.31 & 62.88$\pm$1.29 & 63.16$\pm$1.22 & 79.01$\pm$1.03 & 78.50$\pm$1.05 & 74.69$\pm$1.12 & 75.16$\pm$0.99 \\
  & OURS & $X$, $A$ & \textbf{90.48$\pm$0.78} & \textbf{90.51$\pm$0.81} & \textbf{64.84$\pm$0.59} & \textbf{65.14$\pm$0.47} & \textbf{80.71$\pm$1.05} & \textbf{80.27$\pm$1.36} & \textbf{76.50$\pm$1.22} & \textbf{76.75$\pm$1.22} \\
\bottomrule
\end{tabular}
}
\end{table*}

\begin{table*}[htbp]
\centering
\caption{Node clustering results for unsupervised models in NMI and ARI, and similarity search results in Sim@5, 10, 20 (\%).}
\resizebox{.99\textwidth}{!}{
\label{tab:unsupervised_results}
\begin{tabular}{lcccccccccccccccccccc}
\toprule
\multirow{3}{*}{Method} & 
\multicolumn{5}{c}{ACM} &
\multicolumn{5}{c}{IMDB} &
\multicolumn{5}{c}{DBLP} & 
\multicolumn{5}{c}{Amazon} \\
\cmidrule(lr){2-6}\cmidrule(lr){7-11}\cmidrule(lr){12-16}\cmidrule(lr){17-21} & \multirow{2}{*}{NMI} & \multirow{2}{*}{ARI} & \multicolumn{3}{c}{Sim@} & \multirow{2}{*}{NMI} & \multirow{2}{*}{ARI} & \multicolumn{3}{c}{Sim@}& \multirow{2}{*}{NMI} & \multirow{2}{*}{ARI} & \multicolumn{3}{c}{Sim@}& \multirow{2}{*}{NMI} & \multirow{2}{*}{ARI} & \multicolumn{3}{c}{Sim@}\\
\cmidrule(lr){4-6}\cmidrule(lr){9-11}\cmidrule(lr){14-16}\cmidrule(lr){19-21}  
 & & & 5 & 10 & 20 &  & & 5 & 10 & 20 &  & & 5 & 10 & 20  &  & &5 & 10 & 20 \\
\midrule
  Raw features & 25.82 & 25.10 & 71.80 & 69.20 & 66.16 & 9.11 & 8.80 & 51.67 & 50.52 & 49.04 & 28.92 & 20.52 & 63.91 & 61.25 & 58.51 & 5.61 & 2.22 & 81.11 & 77.69 & 72.38 \\
  Deepwalk  & 31.10 & 32.12 & 71.01 & 70.22 & 69.81 & 11.72 & 10.34 & 49.00 & 48.73 & 48.22 & 34.83 & 30.15 & 62.91 & 62.10 & 61.57 & 8.31 & 4.67 & 72.61 & 71.08 & 70.19 \\
  Node2vec  & 30.92 & 31.11 & 71.01 & 70.21 & 69.80 & 12.32 & 11.27 & 48.71 & 48.20 & 47.91 & 38.32 & 36.54 & 62.89 & 61.93 & 61.02 & 7.45 & 4.13 & 73.83 & 72.68 & 70.79 \\
\midrule
  DGI & 64.83 & 69.07 & 87.92 & 87.63 & 87.00 & 15.62 & 13.26 & 55.51 & 55.41 & 54.57 & 47.21 & 45.09 & 73.26 & 73.22 & 73.16 & 0.30 & 0.31 & 50.62 & 49.17 & 47.45 \\
  GRACE & 60.81 & 67.74 & 86.32 & 85.70 & 85.02 & 17.90 & 16.14 & 58.39 & 58.44 & 57.62 & 44.62 & 41.82 & 77.70 & 77.08 & 76.09 & 21.65 & 13.39 & 75.86 & 74.82 & 73.63 \\
  MVGRL & 69.16 & 74.56 & 88.64 & 88.59 & 88.24 & 15.97 & 17.98 & 58.22 & 56.96 & 55.91 & 53.79 & 49.25 & 77.97 & 77.86 & 77.51 & 0.28 & 0.43 & 48.45 & 46.76 & 45.19 \\
  HDGI & 66.37 & 68.12 & 88.55 & 88.00 & 87.72 & 18.81 & 17.47 & 55.88 & 55.36 & 54.54 & 54.66 & 53.07 & 77.77 & 76.96 & 76.45 & 0.93 & 0.73 & 45.03 & 43.58 & 42.24 \\
  DMGI & 51.80 & 48.39 & 88.58 & 88.31 & 87.96 & 20.37 & 20.76 & 59.69 & 59.02 & 57.80 & 55.49 & 54.34 & \textbf{79.75} & \textbf{78.90} & \textbf{78.25} & 22.75 & 20.30 & 81.69 & 79.80 & 77.97 \\
\midrule
  OURS & \textbf{72.26} & \textbf{75.76} & \textbf{90.27} & \textbf{90.17} & \textbf{89.72} & \textbf{20.58} & \textbf{20.87} & \textbf{61.60} & \textbf{60.87} & \textbf{59.94} & \textbf{56.32} & \textbf{54.75} & 78.11 & 77.41 & 76.74 & \textbf{43.85} & \textbf{41.26} & \textbf{82.45} & \textbf{81.05} & \textbf{79.36} \\
\bottomrule
\end{tabular}
}
\end{table*}
\par
Our method can be grouped into unsupervised methods for heterogeneous graphs. For methods designed for homogeneous graph, we obtain the final embedding by computing the average of node embeddings learned from each graph, i.e. $\bm{H}=\frac{1}{|\mathcal{M}|}\sum_{m\in\mathcal{M}}\bm{H}^{m}$. 
\subsubsection{Parameter Settings}
In our experiments, we closely follow the same dataset split ratio in DMGI \cite{park2020unsupervised} and optimize hyper-parameters for all baselines. 
To make fair comparisons with the most relevant baseline methods, we used the same number of labeled data for ACM and IMDB data sets as in DMGI \cite{park2020unsupervised}. For each class in the rest of data sets, the number of labeled data is set as 20.
All methods use the following training strategy, including the same random seeds for initialization, and the same early stopping on validation set. We evaluated performance on the test set over 10 random parameter initializations.
In addition, the same 64-dimension of representations, 0.5 dropout rate and ReLU activation are used for all baselines to ensure a fair comparison. The learning rate is sets as 0.001. The optimal regularization with weight decay and the number of layers are obtained by grid search for each method. 
We implemented our model using the Adam optimizer \cite{kingma2014adam} and the deep learning extension libraries provided by \cite{geoopt,Fey/Lenssen/2019}. For all other methods, we used the source codes published by the authors, and try to tune them to their best performance. Besides, the experimental results for baselines reproduced by ours are consistent with the reported results of representative work DMGI \cite{park2020unsupervised}. 
For our model, we tune the hyper-parameters by grid-search, including dropout rate $p_a,p_e$ in data augmentation and weights $\beta,\lambda,\gamma$ in objective Eq.~(\ref{eq:loss}). 
\par
The optimal hyper-parameters are obtained by grid search, and the ranges of grid search are summarized in Table \ref{tab:search_space}. Except for shallow methods with 0 layer, the number of hidden layers for unsupervised methods is fixed as 1 layer. For supervised methods, the number of layers is tuned in \{1, 2, 3\}. Another different setting between unsupervised and supervised methods is the patience of early stopping, which is 20 and 100 patience epochs, respectively. The same attention head is set as 8 in attention-based methods. The chosen hyper-parameter for learning rate, weight-decay, and weights $\beta$ of $\mathcal{L}_{node}$, $\lambda$ of $\mathcal{L}_{rel}$, $\gamma$ of $\mathcal{L}_{reg}$, and sample size are shown in Table \ref{tab:chosen_hyperparam}. The chosen dropout rate of attribute masking and edge removing in data augmentation for two graph views are shown in Table \ref{tab:drop}.

\begin{figure}
\begin{minipage}{0.45\textwidth}
\makeatletter\def\@captype{table}
\centering
\resizebox{.86\textwidth}{!}{
\begin{tabular}{lc}
\toprule
Hyper-parameters & Range \\
\midrule
Learning rate & 1e-3, 5e-3\\
Weight decay & 1e-4, 5e-4, 1e-3 \\
$\beta$, $\lambda$, $\gamma$ & 1e-3, 1e-2, 1e-1, 1 \\
$p_a$, $p_e$ & 0, 0.1, 0.2, 0.3, 0.4 \\
\bottomrule
\end{tabular}
}
\caption{Grid search space for hyperparameters.}
\label{tab:search_space}
\end{minipage}
\begin{minipage}{0.54\textwidth}
\makeatletter\def\@captype{table}
\resizebox{.99\textwidth}{!}{
\centering
\begin{tabular}{lccccc}
\toprule
Method & $\mathcal{L}_{2-norm}$ & $\gamma$ & $\beta$ & $\lambda$ & Sample\_size\\
\midrule
 ACM  & 1e-4 & 1.0 & 0.01 & 0.01 & 2,000 \\
 IMDB  & 5e-4 & 0.001 & 0.0 & 0.1 & 1,000 \\
 Amazon  & 1e-4 & 0.01 & 0.001 & 0.0 & 1,000 \\
 DBLP  & 5e-4 & 0.001 & 0.001 & 0.01 & 2,000 \\ 
\bottomrule
\end{tabular}
}
\caption{Hyperparameter specifications.}
\label{tab:chosen_hyperparam}
\end{minipage}
\end{figure}



\begin{table}[htbp]\small
    \caption{Chosen dropout rate in data augmentation.}
    \centering
    \begin{tabular}{lcccc}
    \toprule
     \multirow{2}{*}{Dataset} & \multicolumn{2}{c}{Graph View 1} & \multicolumn{2}{c}{Graph View 2} \\
     \cmidrule(lr){2-3} \cmidrule(lr){4-5} & dropadj $p_e$ & maskfeat $p_a$ & dropadj $p'_e$ & maskfeat $p'_a$ \\
    \midrule
     ACM & 0.1 & 0.1 & 0.2 & 0.1 \\
     IMDB & 0.0 & 0.0 & 0.0 & 0.0 \\
     DBLP  & 0.0 & 0.0 & 0.1 & 0.1 \\ 
     Amazon & 0.4 & 0.1 & 0.4 & 0.1 \\
    \bottomrule
    \end{tabular}
    \label{tab:drop}
\end{table}

\subsection{Performance Comparison}
\subsubsection{Node Classification}
We report Macro-F1 and Micro-F1 with standard deviation for node classification on the test set. We follow DMGI that report the test performance when the performance on validation gives the best result. 
For unsupervised methods, we train a logistic regression classifier for 50 runs, following the standard practice in related works \cite{ren2019heterogeneous}. While for supervised methods, classification results are generated by the model itself. As shown in Table \ref{tab:nc_results}, our model achieves the best performance across all benchmarks compared with not only unsupervised methods, but also supervised methods, demonstrating the comprehensive strength of GCL on modeling heterogeneous graph. 
We also observe that some unsupervised methods such as HDGI and DMGI perform better than strong supervised baselines, suggesting that there exists some potential information unexplored from supervised signals.
At the same time, the smaller standard deviation compared with baselines indicates that GCL can generate a more robust node representation, benefiting from comprehensive dual-view contrastive learning.

\subsubsection{Node Clustering}
We conduct the classical unsupervised task node clustering \cite{park2020unsupervised,hassani2020contrastive} to evaluate our method.
For node clustering, we set the number of clusters to the number of ground-truth classes and then cluster the learned representations using K-Means algorithm. We report the average normalized MI (NMI) and adjusted rand index (ARI) for 50 runs. 
The results shown in Table~\ref{tab:unsupervised_results} suggest that our model outperforms strong unsupervised baselines on four datasets in node clustering, by relative improvements of 6.11\% on average NMI and ARI scores. As we can see, the HDGI performs better than DGI in most cases, indicating that the heterogeneous information in graphs provides additional essential information. Furthermore, by the strength of dual-view contrastive learning, our model outperforms MVGRL, while MVGRL performs even worse than other baselines in some cases. It verifies that the appropriate/applicable views of a graph should be carefully chosen and modeled, otherwise an improper view learned from independent single network embedding methods may have a negative effect.
\subsubsection{Similarity Search}
We further conduct similarity search as another classic task \cite{park2020unsupervised} to evaluate the effectiveness of our method in the unsupervised learning setting. For similarity search, we compute the cosine similarity scores of the node embeddings between all pairs of nodes, and for each node, we rank the nodes according to the similarity score. Then, we calculate the ratio of the nodes that belong to the same class within top-5 (Sim@5), top-10 (Sim@10), and top-20 (Sim@20) ranked nodes.
As shown in Table~\ref{tab:unsupervised_results}, our method also achieves strong performance across baselines on the cosine similarity ranking scores \cite{park2020unsupervised}, including top-5 (Sim@5), top-10 (Sim@10) and top-20 (Sim@20). Specifically, our method performs best except on dataset DBLP, where DMGI performs best on DBLP. Besides, as we can see from Table~\ref{tab:unsupervised_results}, when a model takes into account additional graph knowledge such as heterogeneous relations or different graph views, such as HDGI and MVGRL, the performance of the model will be improved compared with homogeneous-based model DGI. And the improvement will vary with different datasets.
The performance on the two unsupervised protocols demonstrates the comprehensive effectiveness of our self-supervised learning design in modeling the rich semantics and complex structure. 

\begin{figure}[htbp]
\centering
\includegraphics[scale=0.33]{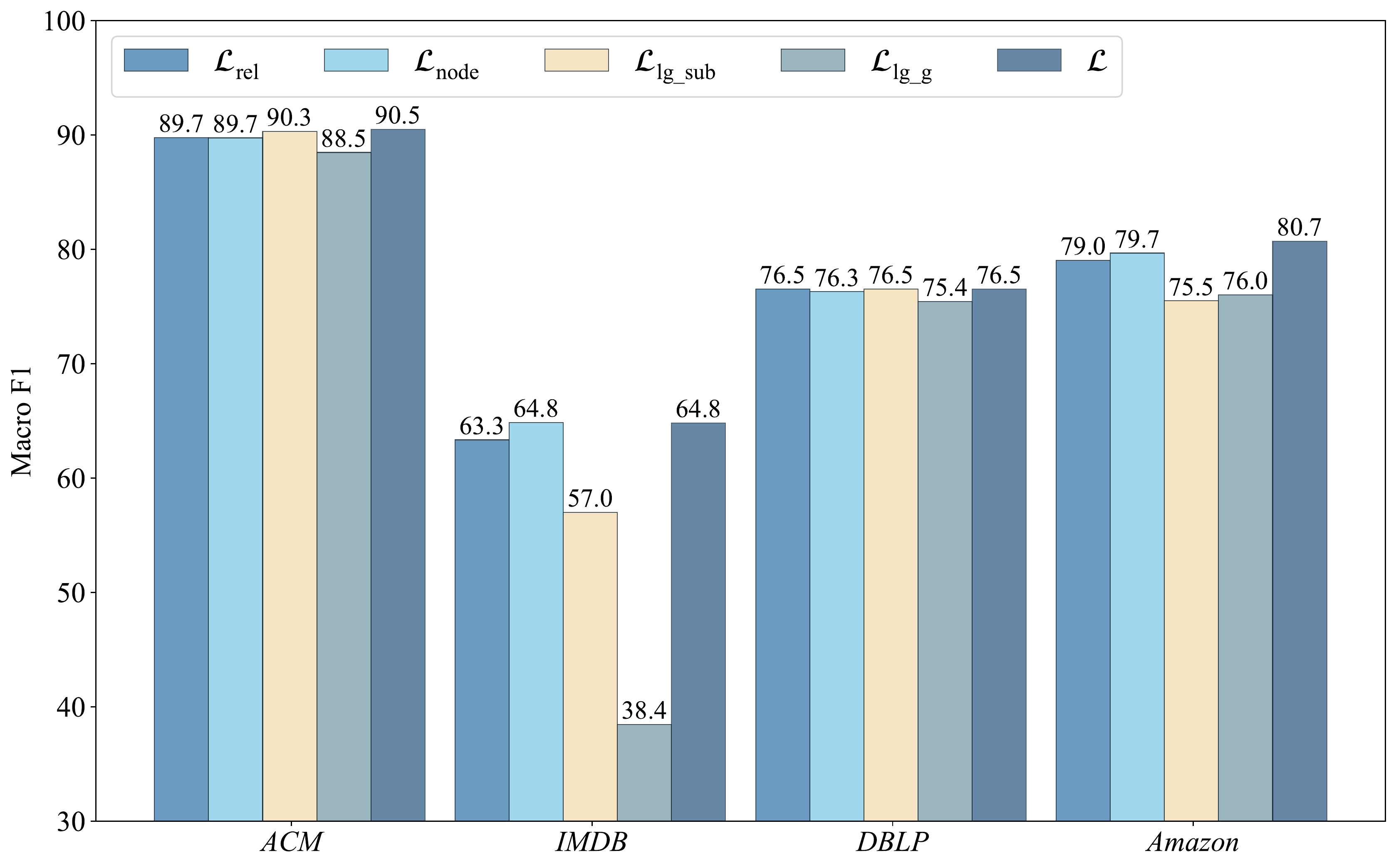}
\caption{Effect of contrastive modes for node classification in Macro-F1 on four datasets.}
\label{fig:contra_mode}
\end{figure}

\begin{table}[htbp]
    \centering
    \caption{Effect of graph views for node classification in Macro-F1 on four datasets.}\label{tab:graph_views}
    \begin{tabular}{lcccc}
    \toprule
     Graph Views & ACM & IMDB & Amazon & DBLP\\
    \midrule
     $wo\_aug$ & 81.93$\pm$2.81 & 64.84$\pm$0.59 & 74.35$\pm$1.48 & 77.81$\pm$1.18 \\
     $geo\_euc$ & 81.82$\pm$1.43 & 57.81$\pm$1.73 & 72.39$\pm$2.06 & 48.10$\pm$1.81 \\
     $geo\_hyp$ & 86.21$\pm$1.00 & 64.50$\pm$0.74 & 74.54$\pm$2.01 & 76.60$\pm$1.90 \\
     $wo\_inter$ & 89.69$\pm$0.77 & 62.83$\pm$0.53 & 75.11$\pm$1.96 & 78.42$\pm$2.15 \\ 
     $wo\_intra$ & 89.70$\pm$0.78 & 62.84$\pm$1.15 & 75.10$\pm$1.96 & 78.97$\pm$1.23 \\ 
     $ensemble$  & \textbf{90.48$\pm$0.78} & \textbf{64.84$\pm$0.59} & \textbf{76.51$\pm$0.95} & \textbf{80.71$\pm$1.05} \\ 
    \bottomrule
    \end{tabular}
\end{table}

\begin{table}[htbp]
    \caption{Effect of contrastive regularization for node classification in Macro-F1, node cluster in NMI and similarity search in Sim@5 on four datasets.}
    \centering
    \begin{tabular}{lcccccc}
    \toprule
     \multirow{2}{*}{Dataset} & \multicolumn{3}{c}{GCL$_{wo\_reg}$} & \multicolumn{3}{c}{GCL} \\
     \cmidrule(lr){2-4} \cmidrule(lr){5-7} & MacF1 & NMI & Sim@5 & MacF1 & NMI & Sim@5 \\
    \midrule
     ACM & 70.84 & 49.52 & 72.43 & 90.48 & 72.26 & 90.27 \\
     IMDB & 49.05 & 16.34 & 55.86 & 64.84 & 20.58 & 61.60 \\
     DBLP  & 72.51 & 40.16 & 75.66 & 80.71 & 56.32 & 78.11 \\ 
     Amazon & 34.93 & 0.16 & 42.69 & 76.51 & 43.85 & 82.45 \\
    \bottomrule
    \end{tabular}
    \label{tab:ab_reg}
\end{table}

\begin{figure*}[tbp]
\centering
\subfigure[DGI]{\includegraphics[width=0.1945\linewidth]{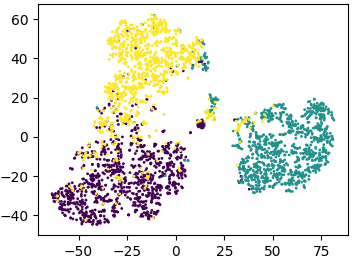}}
\subfigure[MVGRL]{\includegraphics[width=0.1945\linewidth]{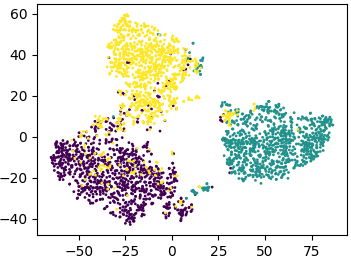}}
\subfigure[HDGI]{\includegraphics[width=0.1945\linewidth]{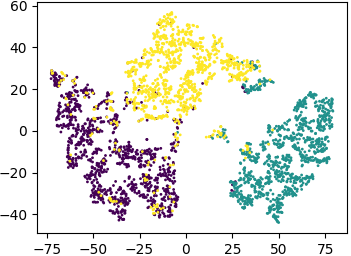}}
\subfigure[DMGI]{\includegraphics[width=0.1945\linewidth]{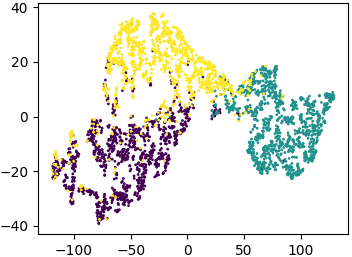}}
\subfigure[GCL]{\includegraphics[width=0.1945\linewidth]{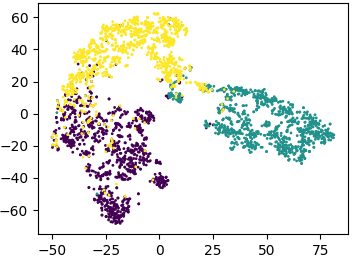}}\\
\subfigure[DGI]{\includegraphics[width=0.1945\linewidth]{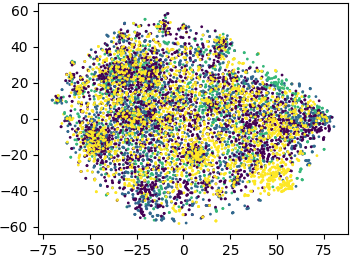}}
\subfigure[MVGRL]{\includegraphics[width=0.1945\linewidth]{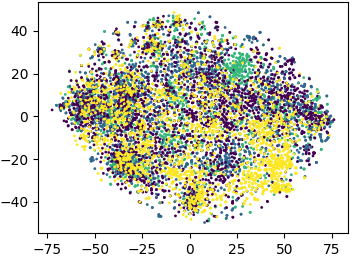}}
\subfigure[HDGI]{\includegraphics[width=0.1945\linewidth]{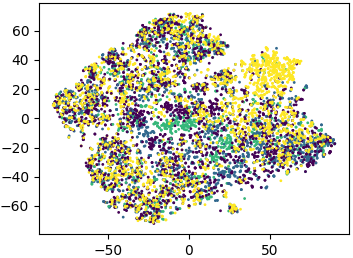}}
\subfigure[DMGI]{\includegraphics[width=0.1945\linewidth]{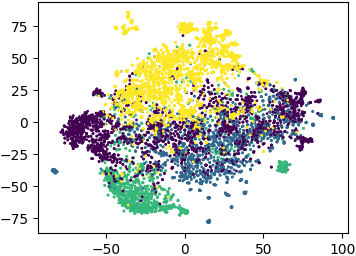}}
\subfigure[GCL]{\includegraphics[width=0.1945\linewidth]{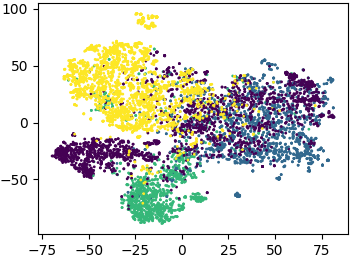}}
\caption{Visualization of the learned node embedding on ACM and Amazon datasets. The first line shows the t-SNE embeddings of the learned nodes from baselines on ACM dataset, and the second line shows the results on Amazon dataset. The color denotes the different node labels. The clusters of the learned GCL model’s embeddings are relatively clearly defined compared with others.}
\label{fig:vis_amazon}
\end{figure*}

\subsection{Experimental Analysis}\label{sec:exp_analysis}
\subsubsection{Effect of Contrastive Modes}
To evaluate the effect of the contrastive modes in our model, we construct four variants by removing one of the contrastive modes.
The variants of removing local-local contrastive $\mathcal{L}_{node}$ defined in Eq.~(\ref{eq:loss_node}) and $\mathcal{L}_{rel}$ defined in Eq.~(\ref{eq:loss_rel}) are denoted as $\neg \mathcal{L}_{node}$, and $\neg \mathcal{L}_{rel}$, respectively.
The variants of removing the first term and the second term of local-global objective defined in Eq.~(\ref{eq:loss_lg}) are denoted as $\neg\mathcal{L}_{lg\_sub}$, $\neg\mathcal{L}_{lg\_g}$, respectively. The proposed model GCL contains above four contrastive modes, denoted as $\mathcal{L}$. 
Each of the variants are re-tuned with the same grid-search as the proposed method.
Figure \ref{fig:contra_mode} illustrates the contribution of each mode on four benchmarks for node classification. Evidently, the ensemble $\mathcal{L}$ consistently perform best across variants. The variants with removing one of the contrastive modes will degrade the performance to some extent.
\par
We further analyze the effect of the contrastive regularization defined in Eq.~(\ref{eq:loss_reg}). In Table \ref{tab:ab_reg}, we show the performance of GCL and its variant GCL$_{wo\_reg}$ of removing the regularization term. We observe that GCL consistently outperforms GCL$_{wo\_reg}$ by a large margin for both supervised and unsupervised tasks, including node classification, node clustering and similarity search, respectively. 
At the same time, the performance of regularization shows that the agreement of meta-path specific node representations is vital and has been effectively imported via contrastive learning.
\subsubsection{Effect of Graph Geometric Views}
To analyze the effect of graph views, we construct two variants by removing data augmentation or geometric encoders. The variant without data augmentation is denoted as $wo\_aug$, utilizing the same graph as the input to different encoders. 
Additionally, we construct two variants $geo\_euc$ and $geo\_hyp$ to show the effects of using two different encoders in graph views.
For variant $geo\_euc$, the two graph encoders are replaced by two conventional Euclidean encoders. 
Another variant uses two hyperbolic encoders in both branches, denoted as $geo\_hyp$.
As shown in Table \ref{tab:graph_views}, the two variants without the enhancement of different geometric encoders not only decrease the overall classification results, but also increase the variance. The results suggest that two different geometric encoders generate effective graph views and have a significant positive impact on the four datasets. 
For data augmentation, we observe that it brings positive effects on most datasets, while it does not make much difference like IMDB. It is consistent with our hypothesis that generating dual-view solely through data augmentation is deficient. 
For instance, we observed that masking node features and removing edges on either views degrade the performance on IMDB dataset, while on other dataset, the data augmentation more or less improves the performance. 
\par
In addition, we investigate the effect of inter-view MI and intra-view MI defined in contrastive objective. The variant of removing inter-view MI in Eq.~(\ref{eq:MI_node_1}), Eq.~(\ref{eq:MI_node_2}) and Eq.~(\ref{eq:MI_rel}) is denoted as $wo\_inter$. Similarly, we have another variant, denoted as $wo\_intra$, by removing the intra-view MI.
As shown in Table \ref{tab:graph_views}, removing each view will decrease the performance. The results validate the effectiveness of graph geometric views from graph views generation and contrastive objective.
\subsubsection{Visualization}
To provide a more intuitive evaluation, we conduct embedding visualization on ACM and Amazon dataset. We plot the learnt node embeddings of DGI, MVGRL, HDGI, DMGI and GCL using t-SNE \cite{van2008visualizing}. The results on ACM and Amazon datasets are shown in Figure \ref{fig:vis_amazon}, in which different colors mean different labels. The visualized representation from the proposed model GCL uses the ultimate node embeddings, which combine the Euclidean and hyperbolic embeddings from different geometric graph views.
For ACM dataset, we can see that nodes colored with green are clearly separated from the other two types of nodes with most methods, such as DGI, MVGRL, HDGI and GCL. However, for the other two types of nodes colored with yellow and purple, the compared methods cannot clearly distinguish between them, where there are some nodes from these two types interspersed with each other. In our proposed GCL, there is less overlap between different categories.
For Amazon dataset with more different relations and fewer labelled data, we can see that DGI, MVGRL and HDGI present blurred boundaries between different types of nodes, because they cannot utilize and fuse all kinds of heterogeneous semantics. For DMGI, there have emerged boundaries between different categories, while nodes are still mixed to some degree. The proposed GCL correctly separates different nodes with relatively clear boundaries. 
To sum up, the proposed method GCL shows the flexible and powerful ability for modeling graphs with multiplex heterogeneous information.

\section{Conclusion}\label{sec:conclusion}
This paper proposed the Geometry Contrastive Learning (GCL) on heterogeneous graphs, a comprehensive self-supervised representation learning method that models rich semantics and complex structure.
Our method derives an expressive view of heterogeneous graph from different geometric spaces, and unifies them under the contrastive learning framework.
The unified contrastive objective is designed to maximize MI between latent representations from inter-view and intra-view at both local-local and local-global semantic levels.
Our method achieves state-of-the-art performance across four benchmarks on node classification, node clustering and similarity search tasks. Extensive ablation studies validate the rationality of the design of GCL in terms of graph geometric views and contrastive modes. We believe that the proposed framework provides a new alternative to expand self-supervised geometry contrastive learning to heterogeneous graph representation learning and enhance downstream applications. 

\begin{acks}
This work was supported in part by the National Key Research and Development Program of China (No. 2021YFB3100500), the NSFC (No. 61872360), the ARC Future Fellowship (No. FT210100097), and the CAS Project for Young Scientists in Basic Research (No. YSBR-008).
\end{acks}

\bibliographystyle{ACM-Reference-Format}
\bibliography{reference}


\end{document}